%% file: neurips_2026.tex
\documentclass{article}

\usepackage[main, final]{neurips_2026} 

\usepackage[utf8]{inputenc} 
\usepackage[T1]{fontenc}    
\usepackage{hyperref}       
\usepackage{url}            
\usepackage{booktabs}       
\usepackage{amsfonts}       
\usepackage{nicefrac}       
\usepackage{microtype}      
\usepackage{xcolor}         
\usepackage{amsmath}
\usepackage{multicol, multirow}
\usepackage{graphicx}
\usepackage{wrapfig}
\usepackage{tabularx}       
\usepackage{algorithm}      
\usepackage{algpseudocode} 
\usepackage[table]{xcolor}
\usepackage{pifont}
\usepackage{caption}
\usepackage{listings}
\usepackage{graphicx}

\title{Mechanistic Circuit Identification for \\ Controllable Data Generation}

\author{
  Nakyung Lee$^{1}$ \quad
  Sangwoo Hong$^{2,*}$ \quad
  Jungwoo Lee$^{1,*}$ \\[0.5em]
  $^{1}$Department of Electrical and Computer Engineering, Seoul National University \\
  $^{2}$Department of Computer Science and Engineering, Konkuk University \\
  \texttt{leena@cml.snu.ac.kr} \quad
  \texttt{swhong06@konkuk.ac.kr} \quad
  \texttt{junglee@snu.ac.kr} \\
  $^{*}$Corresponding authors
}

\begin{document}

\maketitle

\begin{abstract}
While recent advances in data synthesis aim to curate high-quality datasets, most generation pipelines still rely on heuristic prompt-based control. This black-box paradigm provides limited insight into how individual samples interact with a model's underlying learning dynamics. To bridge this gap, we propose a circuit-grounded framework that connects training-dynamics-based data valuation with mechanistic interpretability (MI). 
Specifically, we conceptualize data quality along three complementary utility axes, learnability, challenge, and alignment. First, we uncover specialized model-internal circuits that causally govern these utility signals. Then, moving beyond heuristic prompting toward mechanistic control, we leverage these circuits as controllable interfaces, actively steering generation to produce utility-targeted data. Building on this capability, we introduce SAMS (Stage-Aware Mechanistic Scheduling), which schedules circuit-steered data according to the model's evolving optimization needs. 
Experiments on multiple-choice QA tasks demonstrate that our approach yields precisely controlled data with greater diversity than prompt-based baselines, consistently improving downstream performance and calibration. Ultimately, this work establishes a principled white-box paradigm for interpretable data generation, pioneering the use of MI not just as an analytical tool, but as a practical, controllable interface. Our code is available at \url{https://github.com/nakyungLee20/circuit_datagen}.
\end{abstract}

\section{Introduction}
Data quality is a primary determinant of language model performance, which has led to growing interest in data synthesis, synthetic data generation, and automatic data selection~\cite{datagensurvey,datadiversityit,demystifying,pmpdatasel}. However, most existing pipelines remain largely heuristic and black-box. They are typically judged by downstream gains alone, while offering limited explanation of why some training samples help learning and others do not~\cite{mates,taledd,less}. In language modeling, conventional data quality indicators such as grammaticality, formatting consistency, or self-generated feedback are inherently static and surface-level~\cite{diversitysyn,syndatamultihop}. These heuristic proxies fail to capture the intrinsic sample attribution, how a specific data point shapes the model's parameter updates and contributes to capability formation. While recent gradient-diversity and data influence approaches attempt to address this gap~\cite{datainflu,prismaticsynthesis}, they collapse complex sample impacts into a single scalar magnitude. This abstraction can obscure which internal mechanisms drive useful updates and lead to over-optimizing a single aspect of synthetic data, neglecting the multi-dimensional nature of data quality~\cite{d2pruning,dataselQDC,propella1}.

Recent work has suggested that useful training data cannot be characterized adequately by a single quality score, but should instead be analyzed through complementary notions of whether a sample is \emph{learnable, worth learning, and not yet learnt}~\cite{data3axis}. Motivated by this multi-dimensional view, we formulate data quality as a multi-axis \emph{training utility} profile consisting of \textbf{Learnability} (learnable), \textbf{Challenge} (not yet learnt), and \textbf{Alignment} (worth learning). 
We instantiate these axes using three well-studied training-dynamics-based utility proxies, namely Area Under the Margin (AUM)~\cite{aum}, Error L2-Norm (EL2N)~\cite{el2n}, and Gradient Alignment (GradAlign)~\cite{gradalign}. AUM reflects learnability by measuring whether a sample is learned consistently with a stable margin, EL2N captures informative challenge by identifying samples that induce substantial optimization pressure, and GradAlign measures whether such pressure is aligned with the target objective. Together, these signals distinguish samples that contribute productively to learning from those that are merely hard or noisy.

While such a multi-axis profile provides a richer characterization of data quality than a single score, it remains primarily descriptive. These signals can rank or cluster samples according to their behavior, but they do not explain how such behavior is implemented inside the model. In other words, they tell us which samples appear learnable, challenging, or aligned, but not which internal computations make them so. This motivates our first research question: \textbf{Q1: Are there specific internal computational circuits that govern training-utility signals such as learnability, challenge, and alignment?}

We posit that if high- and low-utility samples consistently drive different learning behaviors, these differences should be reflected in distinct computational pathways within the model. Mechanistic interpretability (MI) offers a natural tool for testing this hypothesis by localizing such pathways as circuits composed of nodes and edges. In this view, training-dynamics metrics quantify how useful a sample is, while MI reveals how that utility is realized internally. If such circuits can be identified and causally validated, they may provide more than post-hoc explanations. By turning utility-associated circuits into actionable interfaces, we can move from retrospectively analyzing data quality toward actively controlling synthetic data generation. This leads to our second research question: \textbf{Q2: Can these discovered circuits serve as causal, circuit-level interfaces for synthetic data generation?}

Based on this motivation, we propose a circuit-grounded framework for utility-driven synthetic data generation. We first discover model-internal circuits associated with established metrics (AUM, EL2N, and GradAlign) on the MCQA task. We then show that these circuits exhibit distinct mechanistic roles aligned with their corresponding utility axes. Building on these findings, we steer the discovered circuits to generate targeted training samples with controllable utility profiles. Finally, to maximize the efficacy of the generated data, we introduce a stage-aware scheduling strategy that modulates their mixture over training time to better match stage-specific learning needs.

\begin{figure}[t]
  \centering
  \includegraphics[width=\columnwidth]{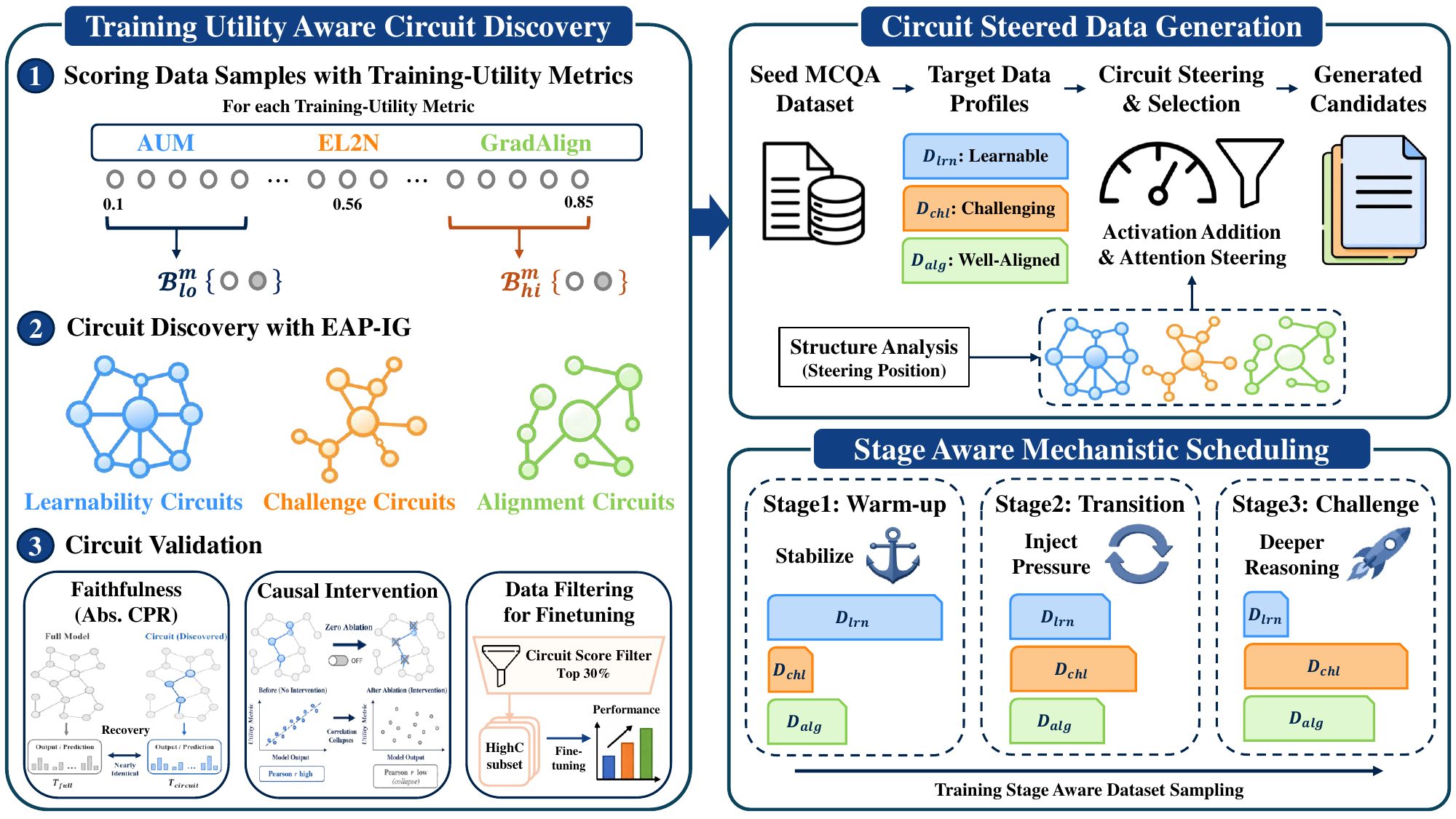}
  \caption{Overview of the proposed framework. We first discover and causally validate internal circuits tied to specific training-utility metrics (Section~\ref{overall_discovery}). We then leverage these circuits to steer the generation process, curating distinct data pools with targeted optimization profiles (Section~\ref{overall_data_generation}). Finally, our SAMS framework dynamically schedules these utility-targeted datasets according to the model's evolving learning phases during downstream fine-tuning (Section~\ref{overall_sams}).}
  \label{fig:mainfigure}
\end{figure}

Our core contributions are summarized as follows:
\begin{itemize}
    \item \textbf{Mechanistic Evidence:} We identify distinct circuits associated with AUM, EL2N, and GradAlign, and provide structural evidence that these training-utility signals are functionally differentiated within the model.
    \item \textbf{Causal Controllability:} We demonstrate that the identified utility circuits can be directly steered to generate training samples with desired training-utility profiles, moving beyond superficial prompt-based control. 
    This highlights the potential of MI as causal control handles, not just descriptive artifacts.
    \item \textbf{SAMS Framework:} We introduce \textbf{SAMS} (\textbf{Stage-Aware Mechanistic Scheduling}), a novel framework that integrates circuit-steered generation with stage-aware scheduling and consistently improves downstream performance over prompt-based baselines. 
\end{itemize}

\section{Related Works}
\paragraph{Mechanistic Interpretability.} Recently, MI has become a promising tool for analyzing the internal mechanisms of large language models~\cite{inductionheads,misurvey,eap,mib}. MI facilitates the reverse-engineering of neural networks by uncovering the granular pathways, or circuits, through which information is processed and transformed~\cite{knowledgecircuits,weightsparse,circuitlora}. These circuits are typically identified through causal intervention methods such as attribution patching and its scalable variants, including EAP and EAP-IG~\cite{eap,eapig}. 
However, existing MI studies predominantly focus on the post-hoc analysis of narrowly defined task-specific behaviors, such as Indirect Object Identification (IOI) or syntactic formatting~\cite{ioi,mda,circuitlora,primer}. In contrast, our work extends the MI paradigm in both scope and application by shifting the focus from simple tasks to the broader complexities of data-level training dynamics, and advancing MI from a passive analytical tool into an active causal interface for practical use.

\paragraph{Data Synthesis.} Current research in data synthesis has primarily focused on improving dataset quality through data mixture optimization, diversity- or model-aware selection, and pruning strategies~\cite{prismaticsynthesis,mates,less,d2pruning,selfevolve}. Despite these advances, the data generation itself remains largely driven by surface-level prompting, instruction engineering, or few-shot in-context learning~\cite{wizard,selfevolve,persona,alpaca}. Such approaches treat the generator as a black box, guiding outputs through linguistic instructions without directly controlling the model's internal computation or its alignment with downstream optimization needs. In contrast, our work moves beyond surface-level distribution shaping by directly intervening in the generation process through circuit-level steering.

\section{Training-Utility-Aware Circuit Discovery}
\label{overall_discovery}
Our first goal is to identify model-internal circuits associated with three training-utility, \emph{learnability}, \emph{challenge}, and \emph{alignment}. As summarized in Figure~\ref{fig:mainfigure}, we measure these dimensions at the sample-level using corresponding established metrics. AUM captures \emph{learnability} through consistent training margins on the correct label, EL2N identifies informative \emph{challenge} based on early prediction errors, and GradAlign evaluates \emph{alignment} via validation-relevant gradient similarity. We then discover the corresponding circuits using EAP-IG~\cite{eapig}, a widely adopted algorithm for circuit identification.

\subsection{Circuit Discovery Setup}
\label{circuit_setup}
We defer the formal definitions of the three training-utility metrics to Appendix~\ref{app:discovery_details} and focus here on how these sample-level utility signals are localized into circuits. We conduct our analysis using the Qwen2.5-1.5B-Instruct model~\cite{qwen25tech} on the SciQ multiple-choice QA benchmark~\cite{sciq}. 

For each metric $m \in \{\mathrm{AUM}, \mathrm{EL2N}, \mathrm{GradAlign}\}$, we compute a scalar utility score $s_i^m$ for each training sample $z_i \in D_{\mathrm{train}}$. We then partition the dataset by ranking samples according to $s_i^m$ and selecting the top-$q\%$ and bottom-$q\%$ subsets:
\begin{equation}
\label{eq:buckets}
\mathcal{B}_{\mathrm{hi}}^{m}=\{i : s_i^m \in \mathrm{top}_q\%\},
\qquad \mathcal{B}_{\mathrm{lo}}^{m}=\{i : s_i^m \in \mathrm{bottom}_q\%\}.
\end{equation}
By leveraging both top- and bottom-ranked data, we frame circuit discovery as a contrastive task. This ensures that we do not only recover general task-solving structures, but instead isolate the specific pathways that drive differences in training utility.

To identify these causal structures, we directly adopt the formulation of EAP-IG~\cite{eapig}, which enables the efficient discovery of circuit edges via gradient approximation. For a given input $x_i$ and a differentiable target function $T(x_i)$, EAP-IG estimates the causal effect of each edge $e$ by computing how much the target behavior shifts between corrupted ($a_{e}^{\mathrm{corr}}$) and clean ($a_{e}^{\mathrm{clean}}$) activation states.
\begin{equation}
\label{eq:eapig_main}
a_{e}^{(j)} = a_{e}^{\mathrm{corr}} + \frac{j}{M} \left( a_{e}^{\mathrm{clean}} - a_{e}^{\mathrm{corr}} \right), \quad \mathrm{Attr}_e(x_i)=\left| \left(a_{e}^{\mathrm{clean}} - a_{e}^{\mathrm{corr}}\right)^{\top} \left( \frac{1}{M}\sum_{j=1}^{M}\nabla_{a_{e}^{(j)}} T(x_i) \right) \right|,
\end{equation}
where $M$ is the number of interpolation steps used to approximate the integral. Edges with larger attribution are interpreted as more influential pathways for mediating utility-related computation.

Executing EAP-IG requires two key components, structurally aligned counterfactual pairs and a target function $T(x_i)$. 
For the counterfactual pairs, prior MI studies have relied on token-level substitutions to analyze localized behaviors within a sequence. However, our objective is to capture the utility profile of an \emph{entire training sample}. We therefore use the original sample embedding $\mathcal{E}(z_i)$ as the clean input and inject Gaussian noise to form the corrupted input:
\begin{equation}
\label{eq:clean_corrupted}
x_i^{\mathrm{clean}}=\mathcal{E}(z_i), \qquad
x_i^{\mathrm{corr}}=x_i^{\mathrm{clean}}+\epsilon, \qquad \epsilon \sim \mathcal{N}(0,\sigma^2 I).
\end{equation}
This continuous perturbation establishes valid counterfactuals while preserving the sequence length of the original sample, which is a practical computational prerequisite for executing EAP-IG.

Regarding the target $T(x_i)$, since the original utility metrics are non-differentiable with respect to internal activations, we use the maximum logit margin, a standard attribution target in circuit discovery, defined as $T(x_i)=\log p_{\theta}(y_{\mathrm{true}} \mid x_i)-\max_{y_{\mathrm{false}}} \log p_{\theta}(y_{\mathrm{false}} \mid x_i)$.

Finally, we aggregate these sample-level attributions across our predefined partitions ($\mathcal{B}_{\mathrm{hi}}^{m}$ and $\mathcal{B}_{\mathrm{lo}}^{m}$) and retain the top-$K$ highest attributed edges for each subset. This yields two isolated circuit families for each metric, associated with high- and low-scoring samples, respectively.

\subsection{Mechanistic Circuit Validation}
\label{circuit_validation}

\begin{figure}[t]
  \centering
  \includegraphics[trim=0cm 0cm 0cm 0.8cm, clip, width=0.95\columnwidth]{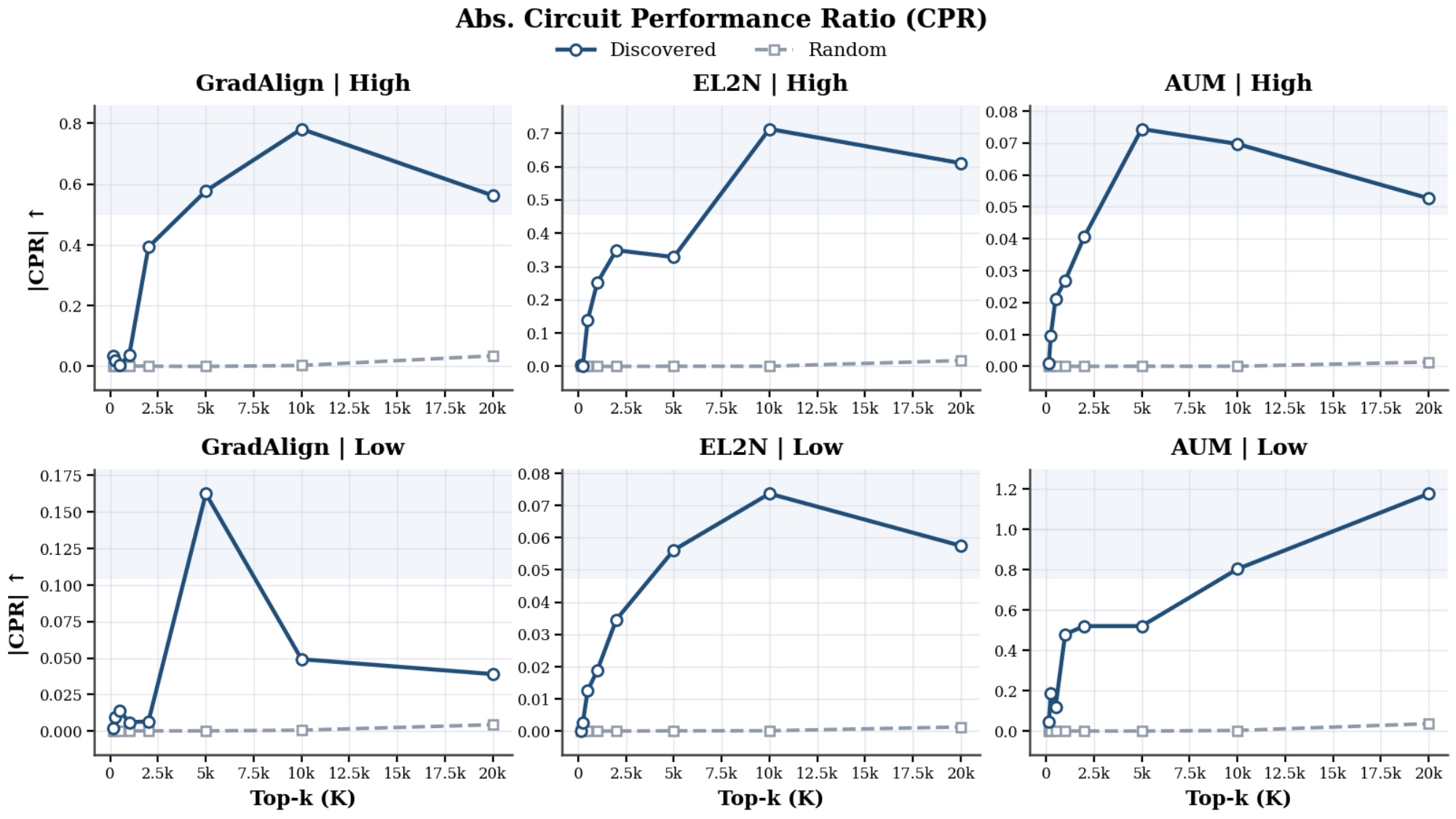}
  \caption{Absolute Circuit Performance Ratio (Abs-CPR) of discovered circuits versus random baselines. The x-axis denotes the number of extracted top-$K$ edges, defining the subgraph size. Higher Abs-CPR indicates that the extracted circuit successfully governs the model's original behavior.}
  \label{fig:abs_cpr}
\end{figure}

\paragraph{Circuit Faithfulness}
To validate our discovered circuits, we evaluate their faithfulness using a variant of the Circuit Performance Ratio (CPR)~\cite{mib}. While standard CPR isolates components that positively restore task performance, training-utility mechanisms can exert both beneficial and detrimental causal forces. 
To capture the \emph{total causal contribution} of a circuit regardless of direction, we adapt this metric into the Absolute Circuit Performance Ratio, defined as Abs-CPR$= |T_{\mathrm{circuit}} - T_{\mathrm{corrupt}}| / |T_{\mathrm{full}} - T_{\mathrm{corrupt}}|$. Here, $T_{\mathrm{full}}$ and $T_{\mathrm{corrupt}}$ are the target responses on clean and corrupted inputs, respectively, and $T_{\mathrm{circuit}}$ is the response when only the top-$K$ EAP-IG-ranked edges process the clean signal. We compare each discovered circuit against random edge subsets of the same size. 
As illustrated in Figure~\ref{fig:abs_cpr}, discovered circuits achieve substantially higher Abs-CPR than random baselines across all three utility metrics, with especially strong recovery for GradAlign-High, EL2N-High, and AUM-Low. As $K$ increases, Abs-CPR generally rises at first, reflecting accumulation of core edges, and then saturates or declines once additional lower-ranked edges introduce less relevant connectivity. This pattern demonstrates that the extracted circuits faithfully concentrate utility-relevant computation within their top-ranked edges, capturing the causal effects far more effectively than arbitrary subgraphs.

\paragraph{Causal Necessity via Circuit Intervention.}
To further demonstrate the causal role of identified circuits, we deactivate them via zero ablation\footnote{For GradAlign, we employ mean ablation instead of zero ablation, since zeroing this circuit can collapse the gradient-alignment signal. A more detailed analysis is provided in Appendix~\ref{app:gr_ga}.}, and then measure how strongly the corresponding sample-level utility scores deteriorate.
Let $q_{\mathrm{full}}(x)$ and $q_{\mathrm{abl}}(x)$ denote the utility score of sample $x$ before and after circuit ablation, respectively. We first quantify the overall score collapse by the average score drop $ S_{\mathrm{drop}}=\mathbb{E}_{x\in B_{\mathrm{all}}}\left[ q_{\mathrm{full}}(x)-q_{\mathrm{abl}}(x) \right]$.
To quantify the loss of discriminative power, we define the high/low score gap under the full model as $G_{\mathrm{full}}= \mathbb{E}_{x\in B_{\mathrm{hi}}}[q_{\mathrm{full}}(x)] - \mathbb{E}_{x\in B_{\mathrm{lo}}}[q_{\mathrm{full}}(x)]$, and analogously the gap after ablation as $G_{\mathrm{abl}} = \mathbb{E}_{x\in B_{\mathrm{hi}}}[q_{\mathrm{abl}}(x)] - \mathbb{E}_{x\in B_{\mathrm{lo}}}[q_{\mathrm{abl}}(x)]$. We then report the Gap Reduction ratio $\mathrm{GapRed}=1-\frac{|G_{\mathrm{abl}}|}{|G_{\mathrm{full}}|}$, where values close to $1$ indicate that the original high/low separation almost completely collapses after deactivating the discovered circuit. 

\begin{wraptable}{r}{0.47\textwidth}
\vspace{-1.2em}
\centering
\caption{Causal intervention of top-250 metric-correlated circuits. Corr.\ denotes the Pearson correlation between $q_{\mathrm{full}}$ and $q_{\mathrm{abl}}$, and GapRed\ measures the collapse of the original high/low score separation after ablation.}
\label{tab:zero_ablation}
\small
\setlength{\tabcolsep}{3pt}
\begin{tabular}{llccc}
\toprule
Metric & Circuit & $S_{\mathrm{drop}}$ & Corr. & GapRed $\uparrow$  \\
\midrule
AUM & High    & 7.042 & -0.007 & \textbf{99.98\%} \\
  & Low     & 7.347 & -0.173 & \textbf{98.05\%} \\
  & Random & 1.171 & 0.940  & 26.58\%    \\
\midrule
EL2N & High    & 0.637 & 0.024  & \textbf{99.53\%}  \\
  & Low     & 0.636 & 0.003  & \textbf{99.58\%}  \\
  & Random & 0.080 & 0.898  & 15.95\%   \\
\midrule
GradAlign & High    & 0.133 & -0.046 & -23.70\%  \\
  & Low     & 0.108 & 0.091  & 19.85\%   \\
  & Random & 0.003 & 0.916  & -61.92\% \\
\bottomrule
\end{tabular}
\vspace{-1.0em}
\end{wraptable}

Table~\ref{tab:zero_ablation} shows that targeted circuit ablation severely disrupts the model's original sample-level utility scoring. Across all metrics, ablation induces substantial score drops ($S_{\mathrm{drop}}$) and destroys rank correlations ($r < 0.1$) compared to random baselines ($r > 0.89$). This severe decorrelation indicates that the discovered circuits act as causal bottlenecks for utility score assignment and sample ranking. For AUM and EL2N, this intervention collapses the high/low score separation (GapRed $>98\%$), validating their discriminative function. Meanwhile, GradAlign exhibits high structural sensitivity, leading to gap fluctuations. Nonetheless, its large score drops and near-zero correlations suggest that the GradAlign circuit serves as both a structurally important processing pathway and a causal bottleneck for gradient-based scoring.

\subsection{Circuit-based Data Filtering}
\label{circuit_filter}

\begin{wraptable}{r}{0.47\textwidth} 
\vspace{-1.5em}
\centering
\caption{Downstream SciQ accuracy on 30\% filtered data subsets. Rand denotes random filtering, and OriS denotes filtering based on the original scalar utility score.}
\label{tab:circuit_filtering}
\setlength{\tabcolsep}{4pt}
\begin{tabular}{lccc}
\toprule
\textbf{Criteria} & \textbf{AUM} & \textbf{EL2N} & \textbf{GradAlign} \\
\midrule
\textbf{HighC} & 85.8 & 86.7 & 86.6 \\
\textbf{LowC}  & 84.8 & \textbf{87.0} & 86.4 \\
\textbf{ContC} & \textbf{86.3} & 85.8 & \textbf{87.7} \\
\midrule
\textbf{Rand}  & 84.0 & 85.2 & 85.8 \\
\textbf{OriS}  & 82.3 & 85.1 & 66.2 \\
\midrule
\textit{Full (100\%)} & \textit{87.4} & \textit{87.6} & \textit{87.4} \\
\bottomrule
\end{tabular}
\vspace{-1.8em}
\end{wraptable}

To bridge our mechanistic insights with practical data curation, we evaluate downstream performance using attribution-guided data filtering. Specifically, we define a circuit-based score and measure the score for each sample using the top-250\footnote{We select this highly sparse subgraph (representing $0.3\%$ of total network edges) because such sparsity is typically preferred to prioritize interpretability and controllability, while still being sufficient to capture the core utility dynamics~\cite{ioi, eap}.} discovered edges. 
Based on these scores, we rank all training samples in the SciQ dataset, retain only the top $30\%$ to form curated subsets, and fine-tune the model on them. Further implementation details are given in Appendix~\ref{app:filtering_details}.

\textbf{Circuit-Based Scoring.} For a sample $x$ and a target circuit $E_t$, we define the circuit score as $S_t(x) = \sum_{e \in E_t} a_e(x) \cdot c_e$. Here, $a_e(x) \in \mathbb{R}$ is a sample-specific scalar activation~\cite{eapig}, and $c_e \in \mathbb{R}$ represents the scalar edge attribution score derived from the initial circuit discovery. Intuitively, $S_t(x)$ quantifies \emph{mechanistic alignment} by measuring how strongly a sample activates the discovered functional circuit. We use this score to construct three filtering criteria by selecting samples with the highest $S_{\mathrm{high}}$ (\textbf{HighC}), highest $S_{\mathrm{low}}$ (\textbf{LowC}), and highest contrastive score $S_{\mathrm{high}} - S_{\mathrm{low}}$ (\textbf{ContC}).

\textbf{Filtering Results and Practical Implications.} Table~\ref{tab:circuit_filtering} demonstrates that circuit-based filtering consistently outperforms random selection and surpasses filtering based on the original scalar metrics. This advantage is particularly pronounced for GradAlign, where raw score filtering severely degrades downstream accuracy while the contrastive circuit signal yields the strongest performance. These results indicate that the circuit-based representation effectively preserves utility-relevant structure that is not fully captured by the scalar score alone.

\subsection{Circuit Structure and Functional Roles}
\label{circuit_structure}
\begin{figure}[t]
  \centering
  \includegraphics[trim=0cm 3.8cm 0cm 3.7cm, clip, width=\columnwidth]{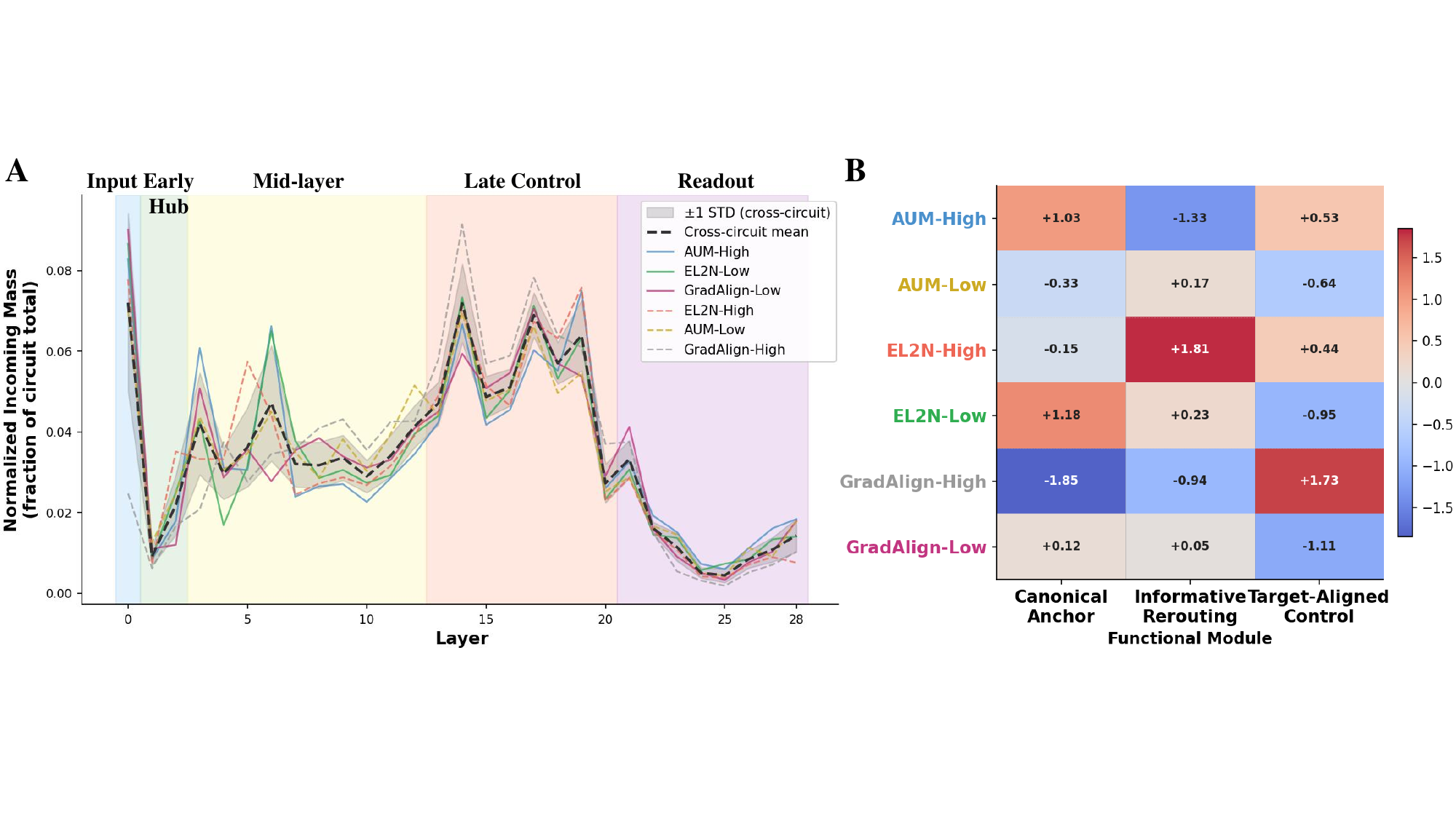}
  \caption{Shared backbone and functional divergence of discovered utility circuits. \textbf{(A)} Layer-wise normalized incoming mass for each circuit. The black dashed line denotes the cross-circuit mean, and the gray band shows $\pm 1$ standard deviation. Narrow variance indicates shared backbone usage, while broader deviations indicate utility-specific routing. \textbf{(B)} Utility-level enrichment of each circuit, reported as cross-circuit $z$-scores. Positive values indicate that a circuit allocates relatively more mass to that module than the cross-circuit average, while negative values indicate relative suppression.}
  \label{fig:circuit_structure}
\end{figure}

To understand how these conceptual utility signals are structurally localized within the model's architecture, we analyze the layer-wise mass distribution of the discovered circuits and examine how their computational mass is routed across different network depths. A detailed node-level breakdown for all circuit profiles is provided in Appendix~\ref{app:structure}.

Our structural analysis reveals a distinct anatomical pattern across the circuits. As shown in Figure~\ref{fig:circuit_structure}A, while all six circuits share a common structural backbone in the earliest and latest layers, substantial divergence emerges in middle layers, where circuits dynamically redistribute their computational mass. To make these routing differences interpretable, we group the discovered edges into three \textit{functional modules} according to their depth and routing type: the \textit{Canonical Anchor} for shallow MLP backbone flow, \textit{Informative Rerouting} for mid-layer Q/K attention shifts, and \textit{Target-Aligned Control} for early-to-late Key-based control. Figure~\ref{fig:circuit_structure}B shows that learnability, challenge, and alignment circuits exhibit distinct module preferences, suggesting that each utility signal is supported by a different structural motif. 

\begin{itemize}
    \item \textbf{Learnability via Canonical Anchor:} Driven predominantly by AUM-High ($+1.03$) and EL2N-Low ($+1.18$), this pathway represents the model's stable default route. Highly learnable, low-optimization-pressure samples do not require complex processing. Instead, they rely on early Key heads (e.g., \texttt{a6<k>}) and layers to directly extract clear label evidence.
    \item \textbf{Challenge via Informative Rerouting:} Dominated by EL2N-High ($+1.81$) and actively suppressed by AUM-High ($-1.33$), this module processes challenging samples located near the decision boundary. For these high-pressure instances, the model bypasses the canonical route, diverting computation through alternative mid-layer attention pathways (e.g., \texttt{a5<k>}) and deeper MLPs (e.g., \texttt{m6}, \texttt{m13}) to resolve ambiguity.
    \item \textbf{Alignment via Target-Aligned Control:} Strongly enriched for GradAlign-High ($+1.73$), this module operates in the later control stages. Rather than relying on shallow shortcuts, it utilizes deeper computation (e.g., \texttt{m13}) to explicitly steer the sample's gradient direction, ensuring the resulting parameter update remains aligned with the target validation objective.
\end{itemize}

\section{Circuit-Steered Data Synthesis}
\label{overall_data_generation}
The preceding results establish that our discovered circuits are faithful, causally relevant, and structurally interpretable. We now pivot from mechanistic interpretation to causal intervention, leveraging these actionable interfaces to generate utility-guided datasets.
Additional steering formulations, implementation details, prompts, and dataset statistics are provided in Appendices~\ref{app:gen_details} and \ref{app:gen_statistics}.

\subsection{Circuit-Steered Data Generation}
\label{circuit_steering}
For circuit-steered generation, we use the same Qwen2.5-1.5B-Instruct model as in circuit discovery and apply two complementary intervention mechanisms. \emph{Activation addition} shifts hidden representations along utility-specific directions, while \emph{targeted attention steering} modulates routing behavior through circuit-relevant attention pathways. 
Concretely, we construct three circuit-steered data pools: $\mathcal{D}_{\text{learnable}}$ prioritizes highly learnable samples with stable margins by activating canonical anchoring pathways. $\mathcal{D}_{\text{challenging}}$ promotes informative difficulty and higher optimization pressure through rerouting-oriented steering. $\mathcal{D}_{\text{aligned}}$ favors target-aligned parameter updates through late-stage control-oriented steering.

\textbf{Activation Addition.} 
Following Representation Engineering (RepE)~\cite{repe}, we construct a steering vector $u_c^{(m)}$ for each module $m \in \mathcal{M}(c)$ of circuit $c$, where $\mathcal{M}(c)$ denotes the set of intervention targets identified in Section~\ref{circuit_structure}. The vector is computed as the mean activation difference between high- and low-utility groups. At each decoding step, we shift the hidden representation $h_t^{(m)}$ via:
\begin{equation}
u_c^{(m)} = \mathbb{E}_{x\sim \mathcal{D}_{c,hi}}[h_\theta^{(m)}(x)] - \mathbb{E}_{x\sim \mathcal{D}_{c,lo}}[h_\theta^{(m)}(x)], 
\quad h_t^{(m)} \leftarrow h_t^{(m)} + \sum_{c:\, m \in \mathcal{M}(c)} \lambda_{c}\, u_c^{(m)},
\end{equation}
where $\lambda_c$ is a steering coefficient dictating circuit amplification ($\lambda_c > 0$) or suppression ($\lambda_c < 0$).

\textbf{Attention Steering.} In addition to shifting hidden representations, we steer sparse attention heads within the discovered circuits. This provides more direct control over token-level routing behavior, complementing activation addition with circuit-targeted modulation of information flow.

As a result, activation addition shapes \emph{what} utility-related circuit state is expressed, while attention steering modulates \emph{how} computation is routed through the discovered pathways. Together, these mechanisms provide causal control at both the representation and routing levels, steering the model's generation toward targeted utility profiles.

\textbf{Selection.} Since circuit steering remains stochastic at generation time, not every candidate faithfully reflects the intended utility profile. We therefore apply a post-generation selection step to remove noisy or weakly matched samples. This improves the consistency of the resulting synthetic pool without altering the steering mechanism itself.

\subsection{Generation Fidelity: Circuit Steering vs. Prompting}
\label{gen_fidelity}
We next evaluate whether circuit-steered generation produces data that is both faithful to the intended training-utility profiles and effective for downstream training. This evaluation proceeds in two stages. First, to isolate the contribution of each functional module, we conduct an internal ablation study that removes each utility axis in turn, assessing the generated data across complexity, quality, and diversity dimensions~\cite{surveyQDC, prismaticsynthesis,dataselQDC}. Second, we compare our circuit-steered datasets against prompt-based baselines under matched target profiles~\cite{ultra,magpie,taledd}, evaluating both their alignment with the original utility metrics and the richness of their learning signals.

\begin{figure}[t]
  \centering
  \begin{minipage}{0.42\textwidth}
    \centering
    \includegraphics[trim=2.5cm 0cm 2.9cm 0cm, clip, width=0.98\linewidth]{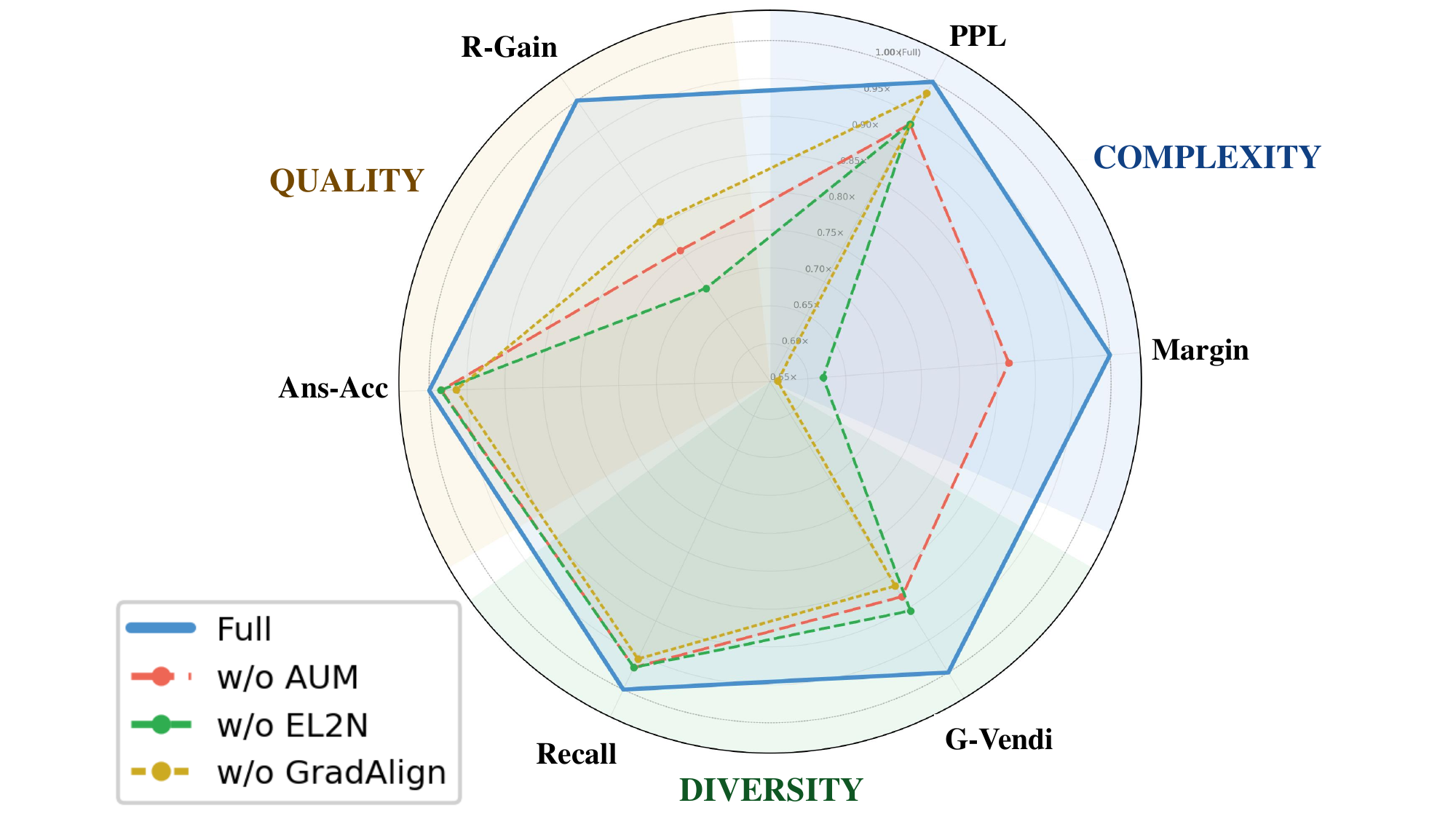}
    \caption{Utility-axis ablation for circuit-steered generation. All metrics are normalized by the Full setting, so values below $1.0$ indicate degradation. See Appendix~\ref{app:gen_fid_details} for detailed metric descriptions.}
    \label{fig:3axisradar}
  \end{minipage}\hfill
  \begin{minipage}{0.56\textwidth}
    \centering
    \includegraphics[trim=0.5cm 0cm 1.2cm 0cm, clip, width=0.96\linewidth]{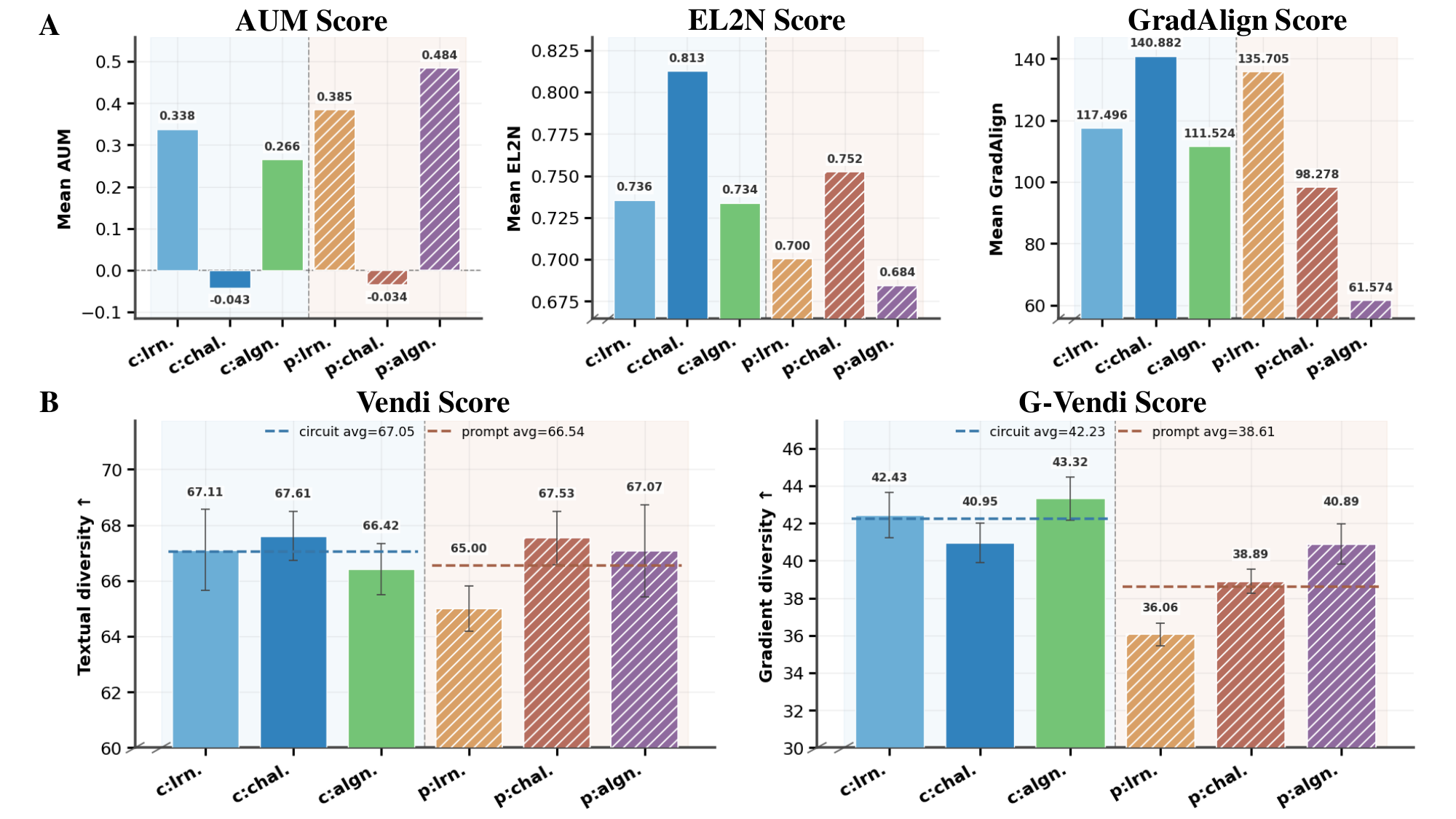}
    \caption{Comparison of circuit-steered (\textbf{c:}) vs. prompt-based (\textbf{p:}) generation. (A) Mean utility scores across learnable, challenging, and aligned target profiles, abbreviated as \textbf{lrn.}, \textbf{chal.}, and \textbf{algn.}. (B) Semantic diversity (Vendi) and gradient-space diversity (G-Vendi).}
    \label{fig:gen_fidelity}
  \end{minipage}
\end{figure} 

\textbf{Necessity of Utility Axes.}
We first investigate the distinct role of each utility-correlated circuit by removing one axis at a time during generation (Figure~\ref{fig:3axisradar}). Without the AUM circuit (\texttt{w/o AUM}), the generated samples maintain fluency (PPL) but suffer a systematic collapse in domain coverage (Recall) and diversity (G-Vendi). Omitting EL2N (\texttt{w/o EL2N}) most notably reduces the additional performance benefit of rationales (R-Gain), indicating a degeneration into trivial samples that lack meaningful reasoning signals. Finally, removing GradAlign (\texttt{w/o GradAlign}) reduces both solving accuracy (Ans-Acc) and the proxy model's confidence margin, demonstrating its role in ensuring target-aligned usefulness. 
Overall, these results suggest that the three axes act synergistically in that AUM ensures learnability (by stabilizing generation), EL2N ensures the presence of not-yet-learnt signals (by preventing trivial mode collapse), and GradAlign ensures the data is worth learning (by preserving target-relevant utility).

\textbf{Faithful Alignment with Utility Profiles.} 
Building on this balanced foundation, we evaluate whether the full circuit-steering framework successfully achieves its intended utility targets. As shown in Figure~\ref{fig:gen_fidelity}A, circuit-steered recipes effectively produce distinct, targeted utility profiles. For instance, $\mathcal{D}_{\text{challenging}}$ yields the highest EL2N alongside a negative AUM score, consistent with generating harder, structurally complex samples. Similarly, $\mathcal{D}_{\text{aligned}}$ achieves a substantially higher GradAlign score than its prompt-based counterpart, indicating a strong preservation of target-relevant optimization signals. 
These results suggest that circuit steering offers a more direct and interpretable interface for controlling training-relevant data properties than prompt-level heuristics.

\textbf{Enhancement of Gradient-Space Diversity.} 
Finally, Figure~\ref{fig:gen_fidelity}B demonstrates that circuit-steered datasets achieve higher diversity, particularly under the G-Vendi metric. This suggests that circuit steering does not merely increase lexical or semantic variation captured by standard Vendi, but broadens the model-induced gradient space, yielding richer optimization signals that are directly tied to downstream robustness~\cite{prismaticsynthesis}.

\section{SAMS: Stage-Aware Mechanistic Scheduling}
\label{overall_sams}
While circuit-steered generation provides independent control over distinct training-utility profiles, treating the resulting pools as a flat mixture ignores the evolving optimization needs of fine-tuning. Prior work establishes that training is highly phase-dependent, with models typically benefiting from stable optimization anchors in early stages before they can effectively absorb high-variance or structurally complex signals~\citep{curri,trhf, el2n}. 
Motivated by this observation, we introduce Stage-Aware Mechanistic Scheduling (\textbf{SAMS}), a stage-wise framework that schedules the circuit-steered pools from Section~\ref{circuit_steering} according to their functional training utility.

Rather than relying on a fixed synthetic mixture, SAMS dynamically reallocates sampling mass across $\mathcal{D}_{\text{learnable}}$, $\mathcal{D}_{\text{challenging}}$, and $\mathcal{D}_{\text{aligned}}$ according to Algorithm~\ref{alg:mechanistic_scheduling}, forming a mechanistically grounded curriculum executed across three progressive phases.
During the \textbf{Warm-up Stage}, training emphasizes $\mathcal{D}_{\text{learnable}}$ to consolidate basic pattern recognition while avoiding excessive gradient variance. In the \textbf{Transition Stage}, we gradually increase the proportion of $\mathcal{D}_{\text{challenging}}$ samples, exposing the model to more structurally demanding signals while preserving a non-trivial anchor component to prevent drift. Finally, in the \textbf{Challenge Stage}, the schedule places greater weight on both $\mathcal{D}_{\text{aligned}}$ and $\mathcal{D}_{\text{challenging}}$, encouraging deeper reasoning and stronger target-aligned behavior.

\begin{wrapfigure}{r}{0.505\textwidth}
\vspace{-0.5em}
\begin{minipage}{\linewidth}
\small
\hrule height 0.8pt
\vspace{2.5pt}
\captionof{algorithm}{Stage-Aware Mechanistic Scheduling}
\label{alg:mechanistic_scheduling}
\hrule height 0.5pt
\vspace{2.0pt}
\begin{algorithmic}[1]
\Require Base $\theta_0$, Steps $T$, Source ratio $r_{\mathrm{src}}$
\Require Source data $\mathcal{D}_{\mathrm{src}}$, Gen. data $\mathcal{D}_{\mathrm{lrn}}, \mathcal{D}_{\mathrm{chl}}, \mathcal{D}_{\mathrm{alg}}$ 
\State Boundaries $t_1, t_2$ ($0 < t_1 < t_2 < T$)
\For{$t = 1$ to $T$}
    \If{$t \le t_1$} \Comment{Warm-up}
        \State $(r_{\mathrm{lrn}}, r_{\mathrm{chl}}, r_{\mathrm{alg}}) \gets (0.60, 0.15, 0.25)$
    \ElsIf{$t_1 < t \le t_2$} \Comment{Transition}
        \State $(r_{\mathrm{lrn}}, r_{\mathrm{chl}}, r_{\mathrm{alg}}) \gets (0.25, 0.45, 0.30)$
    \Else \Comment{Challenge}
        \State $(r_{\mathrm{lrn}}, r_{\mathrm{chl}}, r_{\mathrm{alg}}) \gets (0.10, 0.60, 0.30)$
    \EndIf
    \State Compose $B_t$: Sample $r_{\mathrm{src}}$ from $\mathcal{D}_{\mathrm{src}}$ and $(1 \!-\! r_{\mathrm{src}})$ via $(r_{\mathrm{lrn}}, r_{\mathrm{chl}}, r_{\mathrm{alg}})$
    \State $\theta_t \gets \theta_{t-1} - \eta \nabla_{\theta}\mathcal{L}(B_t; \theta_{t-1})$
\EndFor
\State \textbf{return} $\theta_T$
\end{algorithmic}
\vspace{2pt}
\hrule height 0.8pt
\end{minipage}
\vspace{-1.5em}
\end{wrapfigure}

\paragraph{Downstream Utility via Fine-Tuning}
\label{downstream_perf}
To assess downstream practical utility, we fine-tune a student model (Qwen2.5-0.5B-Instruct~\cite{qwen25tech}) on mixtures of original source data and synthesized data, varying the source-data ratio from 50\% to 70\%. We evaluate both in-domain (SciQ) and out-of-domain (ARC-Easy~\cite{arc}) performance using Accuracy, Expected Calibration Error (ECE)~\cite{ece}, and Negative Log-Likelihood (NLL), where ECE and NLL assess confidence calibration and penalize overconfident mistakes.  

As shown in Table~\ref{tab:ft_results}, training with circuit-steered data consistently outperforms prompt-based generation baselines. On the in-domain SciQ benchmark, SAMS achieves the highest peak accuracy (85.8\%) while substantially reducing overconfidence, as reflected by its lower ECE (0.055). This demonstrates that aligning data complexity with the model's training stage effectively regularizes its internal certainty. The same pattern extends to the out-of-domain ARC-Easy benchmark, where both circuit-steering configurations outperform prompt-tuned models. These results indicate that circuit-steered generation improves downstream utility without sacrificing out-of-domain robustness.

\newcommand{\cmark}{\ding{51}}
\newcommand{\xmark}{\ding{55}}
\begin{table*}[t]
\centering
\caption{Downstream fine-tuning results across source-data ratios. 50\%, 60\%, and 70\% indicate the ratio of original source data to generated data under a fixed total training budget. \textbf{S\&S} indicates circuit-based steering and selection. \textbf{Sch.} applies the exact scheduling ratios from Algorithm~\ref{alg:mechanistic_scheduling} identically to both prompt-generated and circuit-steered pools. We report Accuracy ($\uparrow$), Expected Calibration Error (ECE, $\downarrow$), and Negative Log-Likelihood (NLL, $\downarrow$).}
\label{tab:ft_results}
\resizebox{\textwidth}{!}{
\begin{tabular}{llcc|ccc|ccc|ccc}
\toprule
\multirow{2}{*}{\textbf{Category}} & \multirow{2}{*}{\textbf{Config}} & \multirow{2}{*}{\textbf{S\&S}} & \multirow{2}{*}{\textbf{Sch.}} & \multicolumn{3}{c|}{\textbf{Accuracy ($\uparrow$)}} & \multicolumn{3}{c|}{\textbf{ECE ($\downarrow$)}} & \multicolumn{3}{c}{\textbf{NLL ($\downarrow$)}} \\
\cmidrule(lr){5-7} \cmidrule(lr){8-10} \cmidrule(lr){11-13}
& & & & \textbf{50\%} & \textbf{60\%} & \textbf{70\%} & \textbf{50\%} & \textbf{60\%} & \textbf{70\%} & \textbf{50\%} & \textbf{60\%} & \textbf{70\%} \\
\midrule
\multicolumn{13}{c}{\cellcolor{gray!15}\textbf{In-Domain: SciQ}} \\
\midrule
\rowcolor{gray!5}
Baseline & Source Only (Full) & \xmark & \xmark & \multicolumn{3}{c|}{83.4} & \multicolumn{3}{c|}{0.157} & \multicolumn{3}{c}{1.408} \\
\midrule
\multirow{2}{*}{Prompt Gen.}
& Uniform-Mix      & \xmark & \xmark & 82.9 & 83.7 & 82.8 & 0.070 & 0.078 & 0.093 & 0.541 & 0.551 & 0.588 \\
& Stage-Aware Sch. & \xmark & \cmark & 83.4 & 83.1 & 83.2 & 0.074 & 0.079 & 0.090 & 0.525 & 0.568 & 0.594 \\
\midrule
\cellcolor{white}
& Uniform-Mix      & \cmark & \xmark & 83.9 & 84.5 & \textbf{84.9} & 0.060 & 0.071 & \textbf{0.069} & 0.513 & 0.544 & \textbf{0.540} \\
\rowcolor{gray!10}
\cellcolor{white}\multirow{-2}{*}{Circuit Steering}
& \textbf{SAMS} & \cmark & \cmark & \textbf{84.8} & \textbf{85.8} & 84.4 & \textbf{0.056} & \textbf{0.055} & 0.084 & \textbf{0.491} & \textbf{0.511} & 0.541 \\
\midrule
\multicolumn{13}{c}{\cellcolor{gray!15}\textbf{Out-of-Domain: ARC-Easy}} \\
\midrule
\rowcolor{gray!5}
Baseline & Source Only (Full) & \xmark & \xmark & \multicolumn{3}{c|}{76.2} & \multicolumn{3}{c|}{0.211} & \multicolumn{3}{c}{1.692} \\
\midrule
\multirow{2}{*}{Prompt Gen.}
& Uniform-Mix      & \xmark & \xmark & 72.6 & 72.8 & 72.9 & 0.047 & 0.043 & 0.061 & 0.741 & 0.750 & 0.769 \\
& Stage-Aware Sch. & \xmark & \cmark & 72.4 & 72.1 & 72.4 & 0.060 & 0.052 & 0.072 & 0.768 & 0.772 & 0.793 \\
\midrule
\cellcolor{white}
& Uniform-Mix      & \cmark & \xmark & 73.6 & 73.5 & 74.4 & \textbf{0.037} & \textbf{0.037} & 0.048 & 0.728 & 0.733 & 0.733 \\
\rowcolor{gray!10}
\cellcolor{white}\multirow{-2}{*}{Circuit Steering}
& \textbf{SAMS} & \cmark & \cmark & \textbf{74.1} & \textbf{74.6} & \textbf{74.9} & 0.058 & 0.039 & \textbf{0.047} & \textbf{0.724} & \textbf{0.712} & \textbf{0.717} \\
\bottomrule
\end{tabular}
}
\end{table*}

\section{Ablation Studies}

\paragraph{Isolation of the generation pipeline}
\label{abl_isolation}
To isolate the contributions of steering and selection to downstream performance, we compare their combinations in Table~\ref{tab:isolation}. Steering generally improves downstream performance over unsteered generation, with the largest gain achieved when combined with internal-compatibility filtering. This suggests that steering produces a stronger candidate pool, while filtering removes the remaining low-quality candidates. In contrast, filtering alone provides limited benefit on unsteered generation, suggesting that the two mechanisms are complementary.

\begin{table*}[t]
    \centering
    \caption{Isolation of steering and selection under uniform scheduling.}
    \label{tab:isolation}
    \begin{tabular}{lccc}
        \toprule
        Config (uniform) & SciQ & ECE $\downarrow$ & NII $\downarrow$ \\
        \midrule
        Unsteered + Random
            & $82.80 \pm 0.46$ & $0.077 \pm 0.057$ & $0.487 \pm 0.050$ \\
        Unsteered + Filter
            & $82.27 \pm 0.97$ & $0.057 \pm 0.035$ & $0.507 \pm 0.065$ \\
        Steered + Random
            & $83.10 \pm 0.17$ & $0.057 \pm 0.015$ & $0.480 \pm 0.020$ \\
        Steered + Diversity
            & $83.23 \pm 0.31$ & \textbf{0.050 $\pm$ 0.020} & \textbf{0.477 $\pm$ 0.015} \\
        \textbf{Steered + Filter}
            & \textbf{84.77 $\pm$ 0.38} & $0.063 \pm 0.006$ & $0.510 \pm 0.026$ \\
        \bottomrule
    \end{tabular}
\end{table*}

\paragraph{Sensitivity to phase boundaries}
\label{abl_phase}
Circuit-steered generation produces attribute-enriched pools, while SAMS provides a pool-aware schedule to exploit them. To assess sensitivity to the proposed phase boundaries, we compare SAMS with several alternative scheduling strategies over the same pools. ``Bengio'' denotes an easy-to-hard curriculum based on~\cite{currilearning}.

The results suggest that performance is not highly sensitive to a single set of phase boundaries or mixture ratios. Using the attribute-enriched pools consistently outperforms a generic Bengio curriculum, while SAMS provides the strongest overall performance. This indicates that the key benefit comes from respecting the distinct properties of the steered pools rather than treating them as a homogeneous dataset, while SAMS serves as an effective default pool-aware scheduling strategy.

\begin{table*}[h]
    \centering
    \caption{Sensitivity to phase boundaries and scheduling strategies over the same pools.}
    \label{tab:schedule_sensitivity}
    \resizebox{\textwidth}{!}{%
    \begin{tabular}{lcccc}
        \toprule
        Config & SciQ ACC $\uparrow$ & SciQ ECE $\downarrow$ & SciQ NII $\downarrow$ & ARC (OOD) $\uparrow$ \\
        \midrule
        Bengio (logit-margin as the difficulty proxy)
            & $82.63 \pm 0.23$ & $0.060 \pm 0.026$ & $0.487 \pm 0.051$ & $75.38 \pm 0.34$ \\
        Uniform (33:33:34)
            & $84.77 \pm 0.38$ & $0.063 \pm 0.006$ & $0.510 \pm 0.026$ & $76.09 \pm 0.25$ \\
        Linear interpolation (per-batch)
            & $84.43 \pm 0.91$ & $0.063 \pm 0.015$ & $0.497 \pm 0.042$ & $74.55 \pm 0.99$ \\
        Reversed SAMS (reverse phase order)
            & $84.83 \pm 0.38$ & $0.057 \pm 0.025$ & \textbf{0.483 $\pm$ 0.076} & $75.56 \pm 0.68$ \\
        \textbf{SAMS}
            & \textbf{85.20 $\pm$ 0.56} & \textbf{0.057 $\pm$ 0.023} & $0.490 \pm 0.053$ & \textbf{76.15 $\pm$ 0.30} \\
        \bottomrule
    \end{tabular}%
    }
\end{table*}

\paragraph{Cross-family generalization of the generated data}
\label{abl_generalizability}
Using the same pools and recipe as Table~\ref{tab:zero_ablation}, we evaluate on Llama-3.2-1B-Instruct~\cite{llama} and Gemma-3-1B-it~\cite{gemma} to assess cross-model generalization. Table~\ref{tab:generalization} shows that data generated by steering circuits discovered in Qwen2.5-1.5B-Instruct yields higher SciQ and ARC accuracy than prompt-generated data when used to train student models from two other model families.

\begin{table*}[h]
    \centering
    \caption{Cross-family generalization of the generated data.}
    \label{tab:generalization}
    \small
    \setlength{\tabcolsep}{5pt}
    \begin{tabular}{llcccc}
        \toprule
        Model & Config & SciQ ACC $\uparrow$ & SciQ ECE $\downarrow$ & SciQ NII $\downarrow$ & ARC (OOD) $\uparrow$ \\
        \midrule
        \textbf{Gemma} & Source-only
            & $84.70 \pm 0.00$ & $0.125 \pm 0.007$ & $0.835 \pm 0.064$ & \textbf{75.74 $\pm$ 0.27} \\
        & Prompt-Uniform
            & $84.45 \pm 0.07$ & $0.050 \pm 0.014$ & $0.460 \pm 0.000$ & $74.83 \pm 0.30$ \\
        & Prompt-Sched
            & $84.35 \pm 0.64$ & $0.050 \pm 0.028$ & $0.460 \pm 0.000$ & $73.92 \pm 0.80$ \\
        & Circuit-Uniform
            & $85.10 \pm 0.57$ & $0.055 \pm 0.007$ & $0.480 \pm 0.028$ & $75.55 \pm 0.24$ \\
        & \textbf{SAMS}
            & \textbf{85.75 $\pm$ 0.21} & \textbf{0.045 $\pm$ 0.007} & \textbf{0.425 $\pm$ 0.007} & $75.21 \pm 0.65$ \\
        \midrule
        \textbf{Llama} & Source-only
            & $87.90 \pm 0.00$ & $0.110 \pm 0.000$ & $0.835 \pm 0.035$ & \textbf{79.20 $\pm$ 0.42} \\
        & Prompt-Uniform
            & $86.40 \pm 0.14$ & $0.070 \pm 0.042$ & $0.440 \pm 0.028$ & $77.19 \pm 0.65$ \\
        & Prompt-Sched
            & $84.90 \pm 1.98$ & $0.060 \pm 0.014$ & $0.495 \pm 0.049$ & $76.12 \pm 1.76$ \\
        & Circuit-Uniform
            & $87.50 \pm 0.28$ & \textbf{0.045 $\pm$ 0.007} & \textbf{0.430 $\pm$ 0.014} & $78.98 \pm 0.86$ \\
        & \textbf{SAMS}
            & \textbf{88.15 $\pm$ 0.21} & $0.055 \pm 0.007$ & $0.450 \pm 0.042$ & $79.06 \pm 0.21$ \\
        \bottomrule
    \end{tabular}
\end{table*}

\section{Conclusion}
\label{limitation}
This paper establishes a novel intersection between MI and data synthesis by demonstrating that training data utility can be reverse-engineered into controllable circuit-level variables. Rather than treating data generation as a black-box prompting exercise, our framework actively steers internal modules responsible for learnability, informative challenge, and target alignment to produce data with desired properties. SAMS further shows that these circuit-steered datasets are most effective when their utility profiles are dynamically aligned with the model's evolving optimization needs. Future work will explore extending these localized interventions to a broader range of architectures and tasks, ultimately paving the way for fully white-box, mechanistically guided data curation pipelines.

\section*{Acknowledgements}
This work is in part supported by the National Research Foundation of Korea (NRF, RS-2024-00451435(20\%), RS-2024-00413957(20\%)), Institute of Information \& Communications Technology Planning \& Evaluation (IITP, RS-2025-02305453(10\%), RS-2025-02273157(10\%), RS-2025-25442149(10\%), RS-2021-II211343(10\%), IITP-2026-RS-2026-25615817(10\%)) grants funded by the Ministry of Science and ICT (MSIT), the Culture, Sports and Tourism R\&D Program, funded by the Ministry of Culture, Sports and Tourism, through the Korea Culture Technology Planning and Evaluation Institute, an affiliated institute of the Korea Creative Content Agency in 2026 (RS-2026-2552393(10\%)), Institute of New Media and Communications (INMAC), the BK21 FOUR program of the Education, Artificial Intelligence Graduate School Program (Seoul National University), and Research Program for Future ICT Pioneers, Seoul National University in 2026.

\bibliography{reference}
\bibliographystyle{abbrv}


\appendix
\newpage

\section{Detailed Experimental Setup for Circuit Discovery}
\label{app:discovery_details}

\paragraph{Training-Utility Score Formulation.}
To evaluate training utility, we adopt the metrics of Area Under the Margin (AUM) \citep{aum}, Error L2-Norm (EL2N) \citep{el2n}, and GradAlign \citep{gradalign}. Formally, for a given training sample $x_i$ and its ground-truth label $y_i$, these scores are defined as follows:
\begin{equation}
\begin{aligned}
\label{app-eq:data_metrics}
\mathrm{AUM}(x_i)
&= \frac{1}{T}\sum_{t=1}^{T} \left( z^{(t)}_{y_i} - \max_{j \neq y_i} z^{(t)}_{j} \right), \\
\mathrm{EL2N}(x_i)
&= \left\| p_{\theta_{t_0}}(x_i) - \boldsymbol{e}_{y_i} \right\|_2, \\
\mathrm{GradAlign}(x_i)
&= \big\langle \nabla_{\theta}\ell(x_i, y_i), \, \nabla_{\theta}\mathcal{L}_{\mathrm{val}} \big\rangle.
\end{aligned}
\end{equation}
Here, $z^{(t)}_j$ denotes the logit of class $j$ at training step $t$, $T$ is the total number of recorded steps used to compute AUM, and $p_{\theta_{t_0}}(x_i)$ is the model's predicted probability distribution at an early training checkpoint $t_0$. $\boldsymbol{e}_{y_i}$ represents the one-hot encoded target vector, $\ell(x_i, y_i)$ is the per-sample training loss, and $\mathcal{L}_{\mathrm{val}}$ denotes the loss evaluated on the SciQ validation set. 

We employ \textsc{Qwen2.5-1.5B-Instruct} as our base model and the SciQ training split ($N=11{,}679$) for fine-tuning. This specific pairing is chosen because the model exhibits non-trivial reasoning on SciQ (achieving $50\text{--}60\%$ zero-shot accuracy) while its parameter scale ensures that iterative circuit discovery and validation remain computationally tractable. To evaluate training utility, we dynamically collect three sample-level scores (AUM, EL2N, and GradAlign) for each sample. The model is fine-tuned for 5 epochs using an RTX-Pro 6000 GPU, with EL2N specifically tracked over the first two epochs to capture critical early training dynamics.

\paragraph{Dataset and Contrastive Bucket Curation.}
To facilitate circuit inspection via EAP-IG, we curate contrastive subsets from the scored dataset by isolating the top 15\% and bottom 15\% of samples (1,751 samples each) based on their respective utility metrics. This forms distinct \textit{High-Utility} and \textit{Low-Utility} buckets. We deliberately employ this extreme-percentile binary partitioning to \textbf{maximize the contrastive signal required for attribution patching}. By establishing a high signal-to-noise contrast, this setting cleanly isolates the specific computational subgraphs responsible for desirable versus undesirable model behaviors.
While extending this partition to a fine-grained spectrum (e.g., high/mid/low) remains a valuable avenue for future work, the current binary contrast provides a rigorous foundation for extracting critical circuits without signal dilution. 

\definecolor{codegray}{rgb}{0.5,0.5,0.5}
\definecolor{backcolour}{rgb}{0.97,0.97,0.97}
\definecolor{stringcolor}{rgb}{0.12, 0.47, 0.87} 
\lstdefinelanguage{json}{
    basicstyle=\fontsize{8}{10}\ttfamily, 
    backgroundcolor=\color{backcolour},
    showstringspaces=false,
    breaklines=true,
    frame=lines, 
    numbers=left,
    numberstyle=\tiny\color{codegray},
    stringstyle=\color{stringcolor},
    keywordstyle=\bfseries,
    literate=
     *{:}{{{\color{black}{:}}}}{1}
      {,}{{{\color{black}{,}}}}{1}
      {\{}{{{\color{black}{\{}}}}{1}
      {\}}{{{\color{black}{\}}}}}{1}
      {[}{{{\color{black}{[}}}}{1}
      {]}{{{\color{black}{]}}}}{1},
}
\begin{figure}[h] 
\begin{lstlisting}[language=json]
{
  "clean": "<|im_start|>system\nYou are a helpful multiple-choice assistant.
            Reply with only the correct option letter.<|im_end|>\n
            <|im_start|>user\nQuestion: Long distance runners try to
            maintain constant velocity with very little acceleration or
            deceleration to conserve what?\n\n
            A. pressure\nB. momentum\nC. fuel\nD. energy\n\n
            Return only the correct option letter.<|im_end|>\n
            <|im_start|>assistant\n",
  "label": 35,
  "hf_id": "sciq/train/9939",
  "task": "sciq",
  "score_key": "el2n",
  "bucket": "high",
  "gold_label": "D",
  "label_space": ["A", "B", "C", "D"],
  "corrupt": {"type": "gaussian", "sigma": 0.1, "seed": 2555509},
  "token_len": 252,
  "max_length": 512
}
\end{lstlisting}
\caption{Example of a curated data pair configuration for EAP-IG. The setup explicitly defines the clean prompt, label space, and the continuous corruption strategy.}
\label{app-fig:data_config}
\end{figure}
For each instance within these buckets, we construct a paired setup consisting of a clean input and a corrupted counterpart. As illustrated in Figure~\ref{app-fig:data_config}, the clean input utilizes a chat template, while the corrupted state is generated by injecting continuous Gaussian noise ($\sigma=0.1$) into the embedding space. This ensures our circuit discovery strictly evaluates the pathways responsible for predicting the exact target token.

\paragraph{EAP-IG Implementation Details.}
We implement circuit discovery using EAP-IG, building on its public implementation~\citep{eapig_code}.
The original implementation assumes standard multi-head attention, where query, key, and value heads are aligned one-to-one. However, Qwen2.5 uses grouped-query attention (GQA), in which multiple query heads share a smaller number of key/value heads. To correctly trace edge-level information flow under this architecture, we extend the graph construction and attribution routines to support GQA-style head sharing, referencing prior implementations~\citep{circuit_stability_code}. Specifically, we adapt the graph representation to explicitly track the number of KV heads and accurately route the backward-pass gradients from multiple query heads to their shared KV equivalents.

Given the utility-specific high/low buckets described above, we run EAP-IG with $M=5$ interpolation steps in Eq.~\eqref{eq:eapig_main}, following the recommendation of \cite{eapig}. Attribution is computed at both node and edge granularity. Unless otherwise specified, we rank circuit components by absolute attribution magnitude, which allows us to capture both positively and negatively contributing pathways to the target logit-margin objective.

\begin{table}[h]
\centering
\caption{Sparsity of extracted circuits at varying Top-$K$ edge thresholds. The full computational graph for the Qwen2.5-1.5B model (28 layers, 12 attention heads, 2 GQA groups, and $d_{\text{head}}=128$) yields a total of 365 nodes and 84,715 potential valid edges.}
\label{tab:edge_sparsity}
\vspace{0.3em}
\begin{tabular}{cc}
\toprule
\textbf{Selected Edges (Top-$K$)} & \textbf{Proportion of Total Edges (\%)} \\
\midrule
100 & 0.12\% \\
250 & 0.30\% \\
500 & 0.59\% \\
1,000 & 1.18\% \\
2,000 & 2.36\% \\
5,000 & 5.90\% \\
10,000 & 11.80\% \\
15,000 & 17.71\% \\
\bottomrule
\end{tabular}
\end{table}

\section{Detailed Experimental Setup for Circuit Validation}
\label{app:validation_details}

\paragraph{Zero-Ablation Experiment Setup.} 
As described in Section~\ref{circuit_validation}, we perform zero-ablation experiments to test whether the discovered circuits causally drive the corresponding utility-score behavior. For each utility metric $m \in \{\mathrm{AUM}, \mathrm{EL2N}, \mathrm{GradAlign}\}$, we take the top-$250$ edges discovered by EAP-IG and disable them by replacing their activations with zero during the forward pass. We then repeat the same fine-tuning and utility-score collection procedure described in Appendix~\ref{app:discovery_details} under the ablated model, denoted by $q_{\mathrm{abl}}(x)$.

Because the utility metrics have different mathematical forms, we use two ablated-score collection protocols. For dynamic training-dependent metrics such as AUM and EL2N, we repeat fine-tuning with the corresponding circuit edges persistently ablated, and collect $q_{\mathrm{abl}}(x)$ along this ablated training trajectory in the same manner as $q_{\mathrm{full}}(x)$. In contrast, for gradient-based metrics such as GradAlign, we compute $q_{\mathrm{abl}}(x)$ directly from an ablated forward-backward pass at the same checkpoint used for $q_{\mathrm{full}}(x)$, without additional training. As a control baseline, we randomly sample an equivalent number of edges ($K=250$) and apply the identical zero-ablation procedure. This allows us to distinguish targeted causal effects from score deviations caused by random architectural disruption.

To quantify the ablation impact, we specifically analyze the samples comprising our previously curated high- and low-utility buckets. For these targeted samples, we compute the Pearson correlation coefficient between the utility score rankings produced by the full model $q_{\mathrm{full}}(x)$ and the ablated model $q_{\mathrm{abl}}(x)$. Additionally, we observe the collapse in the score gap between the buckets. A substantial decrease in both correlation and bucket gap robustly indicates that the ablated circuit is causally responsible for generating the specific utility signal.

\paragraph{Denominator instability of GapRed for GradAlign}
\label{app:gr_ga}
The negative GapRed values for GradAlign arise from denominator instability rather than indicating that the discovered circuit is ineffective. GapRed normalizes by the high--low bucket mean gap, which is nearly zero for GradAlign ($\Delta_{\mathrm{full}} = 0.00485$), causing the ratio to explode and $1-\mathrm{ratio}$ to become negative. Unlike AUM and EL2N, which accumulate stable signals over the training trajectory, GradAlign is a checkpoint-dependent static score. As a result, its bucket means are much closer together, making the denominator in GapRed nearly zero and unstable. This interpretation is supported by the fact that even the random-edge ablation yields GapRed $-61\%$ while preserving the ranking. To avoid this denominator instability, we report the more stable diagnostics in Table~\ref{app-tab:gr_score_change}.

\begin{table}[h]
    \centering
    \caption{Effect of circuit ablation on utility scores. $\mathrm{mean\_abs}\Delta$ is the mean absolute change in sample scores after ablation. $E_\Delta(D_H)$ and $E_\Delta(D_L)$ report this quantity within the high- and low-score buckets, respectively. Pearson(full, abl) is the correlation between scores before and after ablation.}
    \label{app-tab:gr_score_change}
    \begin{tabular}{lrrrr}
        \toprule
        Ablated Circuit & $\mathrm{mean\_abs}\Delta$ & $E_\Delta(D_H)$ & $E_\Delta(D_L)$ & Pearson(full, abl) \\
        \midrule
        AUM-High    & $7.042$ & $10.684$ & $3.399$ & $-0.007$ \\
        AUM-Low     & $7.347$ & $11.065$ & $3.628$ & $-0.173$ \\
        AUM-Random  & $1.171$ & $1.240$  & $1.102$ & $0.940$ \\
        \midrule
        EL2N-High   & $0.637$ & $0.404$  & $0.870$ & $0.024$ \\
        EL2N-Low    & $0.636$ & $0.404$  & $0.869$ & $0.003$ \\
        EL2N-Random & $0.080$ & $0.157$  & $0.003$ & $0.898$ \\
        \midrule
        GA-High     & $0.133$ & $0.174$  & $0.092$ & $-0.046$ \\
        GA-Low      & $0.108$ & $0.122$  & $0.093$ & $0.091$ \\
        GA-Random   & $0.003$ & $0.002$  & $0.004$ & $0.916$ \\
        \bottomrule
    \end{tabular}%
\end{table}

The table shows that random-edge ablation produces almost no score change and preserves the original ranking, indicating that the utility signal remains intact. In contrast, ablating the discovered circuits causes substantially larger score changes across samples while simultaneously disrupting the original score ordering. Together, these complementary diagnostics show that the discovered circuits are causally responsible for the GradAlign utility signal rather than introducing only small perturbations.

\paragraph{Data Filtering Experimental Setup.}
\label{app:filtering_details}
To demonstrate the practical applicability of mechanistic insights, we leverage the discovered circuits to perform data filtering prior to our main data generation pipeline. As defined in Section~\ref{circuit_filter}, the circuit score $S_t(x)$ quantifies the mechanistic alignment. For each sample $x$ in the SciQ training split, we compute its circuit score with respect to the discovered high- and low-utility circuits of metric $m$. We then rank all training samples according to the corresponding circuit-based selection score and retain the top $q=30\%$ of examples. Formally, our filtering setups are defined as follows:
\begin{itemize}
    \item \textbf{HighC Filter} (High-quality Circuit-based): $\mathcal{D}_{\text{high}} = \mathop{\mathrm{TopQ}}_{q} \left( S_{\text{high}}^M(x) \right)$
    \item \textbf{LowC Filter} (Low-quality Circuit-based): $\mathcal{D}_{\text{low}} = \mathop{\mathrm{TopQ}}_{q} \left( S_{\text{low}}^M(x) \right)$
    \item \textbf{ContC Filter} (Contrastive Circuit-based): $\mathcal{D}_{\text{cont}} = \mathop{\mathrm{TopQ}}_{q} \left( S_{\text{high}}^M(x) - S_{\text{low}}^M(x) \right)$
\end{itemize}
We compare these mechanistic filtering strategies against three baselines: \textbf{Full} (using the entire dataset), \textbf{Rand Filter} (a randomly selected $q\%$ subset), and \textbf{OriS Filter} (the top $q\%$ selected using the original, un-patched utility formulations such as EL2N). For all downstream validation experiments on these subsets, we fine-tune Qwen2.5-1.5B-Instruct for 3 epochs with a learning rate of $2 \times 10^{-5}$ on the same RTX-Pro 6000 GPU.

\section{Distinctness of the discovered utility circuits}
\label{app:circuit_distinctness}
Our discovery is contrastive by construction: for every utility metric, we contrast high- vs.\ low-scoring buckets, so components shared across all inputs (e.g., general reading comprehension and answer formatting) are mitigated, while utility-specific edges are emphasized. Using extreme utility partitions further strengthens the contrastive signal. This design choice is empirically supported by the Table~2 results, where contrastive circuit signals (ContC) consistently outperform high- or low-only discovery. This partition design is further validated by two additional analyses, confirming that the emphasized circuits do not form a single generic subnetwork.

\paragraph{Structural distinctness.} As shown in Table~\ref{app-tab:circuit_jaccard}, all off-diagonal Jaccard overlaps are below $0.56$, compared with identical circuits. The largest overlap (AUM-hi / EL2N-lo) is expected, since a high-margin example is also a low-error example.

\begin{table*}[h]
    \centering
    \caption{Jaccard overlap between the top-250-edge circuits.}
    \label{app-tab:circuit_jaccard}
    \small
    \begin{tabular}{lrrrrrr}
        \toprule
        & AUM-hi & AUM-lo & EL2N-hi & EL2N-lo & GA-hi & GA-lo \\
        \midrule
        \textbf{AUM-hi}  & 1.000 & 0.355 & 0.374 & 0.553 & 0.319 & 0.458 \\
        \textbf{AUM-lo}  & 0.355 & 1.000 & 0.323 & 0.370 & 0.282 & 0.441 \\
        \textbf{EL2N-hi} & 0.374 & 0.323 & 1.000 & 0.374 & 0.359 & 0.362 \\
        \textbf{EL2N-lo} & 0.553 & 0.370 & 0.374 & 1.000 & 0.292 & 0.393 \\
        \textbf{GA-hi}   & 0.319 & 0.282 & 0.359 & 0.292 & 1.000 & 0.377 \\
        \textbf{GA-lo}   & 0.458 & 0.441 & 0.362 & 0.393 & 0.377 & 1.000 \\
        \bottomrule
    \end{tabular}
\end{table*}

\paragraph{Metric-conditioned functional selectivity.} Using each row circuit's score as a predictor, we report its ROC-AUC for distinguishing the high- and low-score groups defined by each column utility, along with Cohen's $d$ for the difference in mean circuit scores between the two groups. Bold values indicate the strongest positive discrimination within each utility column.

\begin{table*}[h]
    \centering
    \caption{Metric-conditioned functional selectivity of the discovered circuits. Each cell reports ROC-AUC and Cohen's $d$.}
    \label{app-tab:circuit_functional_selectivity}
    \small
    \begin{tabular}{lccc}
        \toprule
        Circuit $\backslash$ Utility Bucket & AUM & EL2N & GradAlign \\
        \midrule
        AUM-high
            & \textbf{$0.648$} ($d=+0.46$)
            & $0.354$ ($d=-0.44$)
            & $0.457$ ($d=-0.13$) \\
        EL2N-high
            & $0.327$ ($d=-0.51$)
            & \textbf{$0.673$} ($d=+0.52$)
            & $0.551$ ($d=+0.15$) \\
        GA-high
            & $0.384$ ($d=-0.32$)
            & $0.606$ ($d=+0.24$)
            & \textbf{$0.569$} ($d=+0.17$) \\
        \bottomrule
    \end{tabular}
\end{table*}

Within the AUM column, the AUM-high circuit shows the strongest positive discrimination (AUC 0.648), whereas within the EL2N column, the EL2N-high circuit performs best (AUC 0.673). The opposing cross-metric responses are consistent with the reversed semantics of AUM and EL2N. If the discovered circuits merely reflected a generic logit-reading mechanism, they would not consistently discriminate the utility partitions associated with each metric. The observed patterns support a metric-conditioned relationship between the discovered circuits and the corresponding utility axes.

\section{Detailed Topology of Training-Utility Circuits}
\label{app:structure}
We provide a finer-grained summary of the information processing pathways for the six discovered circuits corresponding to the high and low regimes of AUM, EL2N, and GradAlign. As detailed in Table~\ref{app-tab:struct_circuits}, the circuits generally share a common scaffold while exhibiting metric-specific divergence. Recurring components include an early input-loading interface through layer-0 value heads, dominant routing through $m1$ and $m2$, and a late pre-output bottleneck around $m23 \rightarrow \mathrm{logits}$. However, the principal differences emerge in the mid-to-late layers, where distinct attention banks, MLP hubs, and suppressive branches selectively dominate different utility axes. 

\begin{table*}[h]
\centering
\small
\caption{Representative location of the six discovered training-utility circuits. ``Key Pathways'' lists the dominant or most discriminative nodes and edges. ``suppr.'' denotes edges or nodes with suppressive(negative) attribution.}
\label{app-tab:struct_circuits}
\begin{tabularx}{\textwidth}{l X}
\toprule
\textbf{Circuit} & \textbf{Key Pathways} \\
\midrule
AUM-high &
\texttt{input $\rightarrow$ a0.h6--11$\langle$v$\rangle$}; 
\texttt{m1,m2}; 
\texttt{m1,m2 $\rightarrow$ a6.h6--11$\langle$k$\rangle$}; 
\texttt{a14}; 
late L12/13/17/18/19; 
\texttt{m23 $\rightarrow$ logits} \\
AUM-low &
\texttt{m1 $\rightarrow$ a2$\langle$q$\rangle$}; 
\texttt{m1,m2 $\rightarrow$ a5$\langle$k$\rangle$}; 
\texttt{m1,m2 $\rightarrow$ m3} (suppr.); 
\texttt{m27 $\rightarrow$ logits} (suppr.) \\
EL2N-low &
\texttt{input $\rightarrow$ a0.h6--11$\langle$v$\rangle$}; 
\texttt{m1 $\rightarrow$ m2 $\rightarrow$ m3}; 
\texttt{m1,m2 $\rightarrow$ a6.h6--11$\langle$k$\rangle$}; 
\texttt{m23 $\rightarrow$ logits} \\
EL2N-high &
\texttt{m1 $\rightarrow$ a2$\langle$k$\rangle$} (suppr.); 
\texttt{m1,m2 $\rightarrow$ a5$\langle$k$\rangle$}; 
\texttt{m4,m6,m13}; 
\texttt{m2 $\rightarrow$ a6,a14} (suppr.) \\
GradAlign-high &
\texttt{m1,m2 $\rightarrow$ m13}; 
\texttt{m1,m2 $\rightarrow$ a14$\langle$k$\rangle$} (suppr.); 
\texttt{m1,m2 $\rightarrow$ a4$\langle$k$\rangle$} (suppr.); 
L13--L20 \\
GradAlign-low &
\texttt{input $\rightarrow$ a0.h6--11$\langle$v$\rangle$}; 
\texttt{m1,m2}; 
\texttt{m1,m2 $\rightarrow$ m3} (suppr.); 
\texttt{m23 $\rightarrow$ logits}; 
\texttt{m27 $\rightarrow$ logits} (suppr.) \\
\bottomrule
\end{tabularx}
\end{table*}

Specifically, the functional characteristics of each circuit can be summarized as follows:
\begin{itemize}
    \item \textbf{AUM Circuits:} The \texttt{AUM-high} circuit acts as an anchor backbone, characterized by stable ingress and canonical dispatch with strong early hubs and direct late readout. In contrast, the \texttt{AUM-low} circuit forms an ambiguity-sensitive branch. It is weaker, more heterogeneous, and resembles a failure-adjacent route rather than a clean canonical path.
    \item \textbf{EL2N Circuits:} The \texttt{EL2N-low} circuit represents a canonical low optimization pressure route, maintaining strong ingress, stable early chaining, and efficient readout. Conversely, the \texttt{EL2N-high} circuit functions as a rerouting-pressure branch, shifting mass away from the canonical route toward alternative mid-layer routing and deeper MLP processing.
    \item \textbf{GradAlign Circuits:} The \texttt{GradAlign-high} circuit operates as a validation-aligned control mechanism, featuring late control entry through \texttt{m13} combined with the suppression of shortcut-like routes. Finally, the \texttt{GradAlign-low} circuit reflects a short-path evidence mode. While it shares the standard ingress and direct readout, it applies strong suppressive shaping around \texttt{m3} and late output nodes.
\end{itemize}
By abstracting these fine-grained structural pathways across different utility metrics, we derive the three overarching functional modules introduced in the main text (Section~\ref{circuit_structure}, Figure~\ref{fig:circuit_structure}B).  

\begin{figure}[h]
  \centering
  \includegraphics[trim=0cm 2.2cm 0cm 2cm, clip, width=\columnwidth]{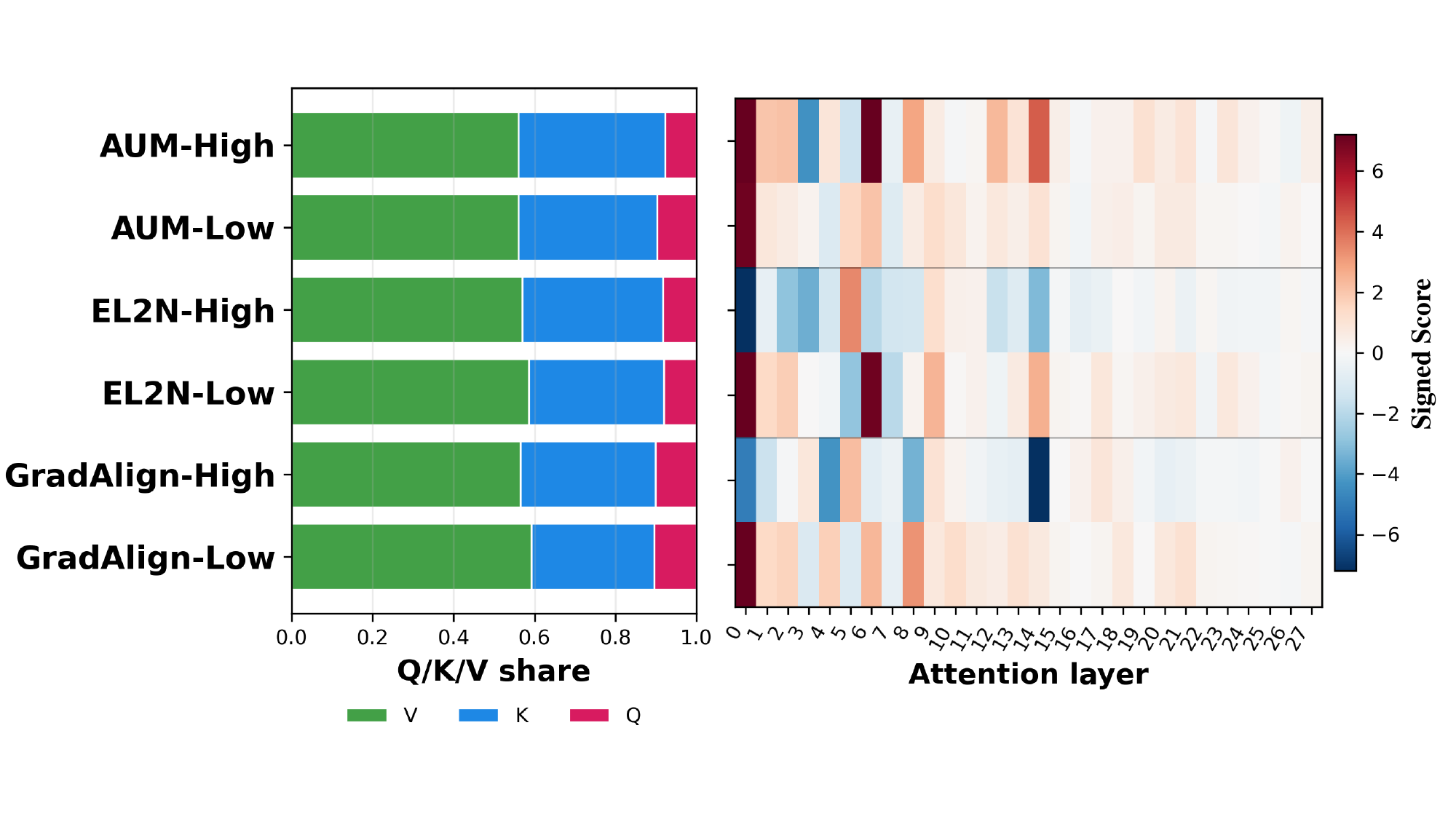}
  \caption{Attention-interface decomposition of SciQ utility circuits.
  Left: Q/K/V share of attention-interface attribution mass across the six discovered circuits.
  Right: signed attention-layer profile, computed by aggregating attention-node attribution scores within each layer.}
  \label{app-fig:attention}
\end{figure}

\paragraph{Attention-interface Decomposition.}
We further analyze the discovered circuits by decomposing their attention-interface mass into query, key, and value pathways, and by aggregating signed attribution across attention layers.
As shown in Figure~\ref{app-fig:attention}, the Q/K/V composition is largely conserved across the six utility circuits. Value pathways dominate the attention-interface mass, followed by key pathways, while query pathways account for a smaller fraction. This suggests that the circuits share a common value-mediated information-loading backbone. 
In contrast, the signed layer-wise profiles reveal clear metric-specific structure. AUM and EL2N-low circuits preserve a positive early attention scaffold, EL2N-high shifts attribution toward alternative mid-layer attention routes, and GradAlign-high exhibits strong suppressive attribution in mid-to-late attention layers.
Thus, the main functional differences are not explained by global Q/K/V proportions alone, but by where signed attention routing is amplified or suppressed across depth. 
This validates our use of \textbf{absolute attribution magnitude} for circuit extraction, as training-utility circuits depend on both supportive activation pathways and suppressive control pathways.

\section{Implementation Details for Circuit-Steered Data Generation}
\label{app:gen_details}
This section provides additional implementation details for the circuit-steered data generation method introduced in Section~\ref{circuit_steering}. Specifically, we elaborate on the targeted attention steering mechanism, the internal-compatibility selection process, and hyperparameters used during generation.

\subsection{Detailed Circuit Steering Implementation}
\paragraph{Attention Steering.}
Activation addition shifts the internal \emph{state} of a circuit, but it does not directly control how information is routed across tokens. We therefore complement activation addition with targeted attention steering. The intervention is applied only to attention heads selected from the discovered circuit presets in Table~\ref{app-tab:circuit_presets}. Let $A_{\ell,h,t,j}$ denote the pre-softmax attention logit from query position $t$ to key position $j$ at layer $\ell$ and head $h$. For a steering profile $r$, we apply a sparse logit perturbation
\begin{equation}
\widetilde{A}_{\ell,h,t,j} = \frac{A_{\ell,h,t,j}}{\tau(s_{r,\ell,h})} + s_{r,\ell,h} \cdot \mathcal{M}(t, j).
\end{equation}
Here, $\mathcal{M}(t, j)$ is a structural routing mask that prioritizes key semantic regions such as the prefix and the local context window around the current decoding step $t$. The bounded temperature function $\tau(\cdot)$ modulates the sharpness of the selected attention head in proportion to the steering scale. The predefined profile-dependent steering strengths ($s_{r,\ell,h}$) are specified by the attention scales in Table~\ref{app-tab:profile-scales}. This intervention remains sparse, localized, and tied to the circuit's functional role, altering routing behavior without indiscriminately disrupting the model's global attention dynamics.

\paragraph{Selection by Internal Compatibility.}

\begin{wrapfigure}{r}{0.45\textwidth}
  \centering
  \vspace{-1.1em}
  \includegraphics[trim=1cm 1cm 1cm 1cm, clip, width=\linewidth]{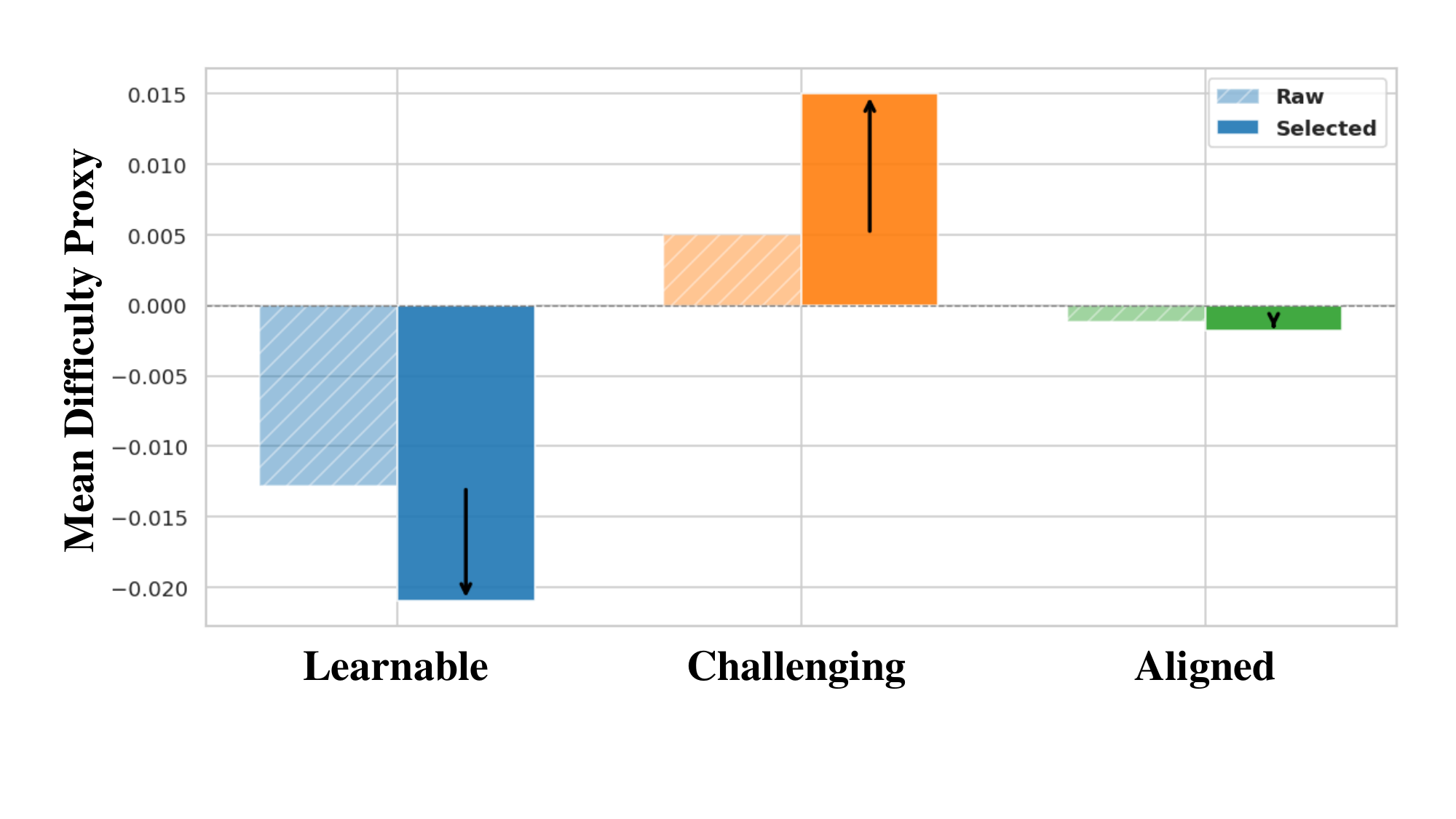}
  \caption{Comparison of mean difficulty proxy between raw and selected datasets.}
  \label{app-fig:selction}
  \vspace{-0.4em}
\end{wrapfigure}

We apply a lightweight selection step that keeps candidates whose likelihood profile is most compatible with the intended diagnostic route. For each source example, we first discard candidates that violate task-specific formatting or parsing constraints. We then rescore each remaining candidate by measuring its teacher-forced log-likelihood under the same target circuit configurations used during steering.

Given an input $x$ and a candidate completion $\tilde{y}$, we compute the average token log-likelihood under each diagnostic route $k$:
\begin{equation}
\ell_k(\tilde{y}\mid x) = \frac{1}{|\tilde{y}|} \sum_{t=1}^{|\tilde{y}|} \log p_{\theta,k}(\tilde{y}_t \mid x,\tilde{y}_{<t}),
\end{equation}
where $k \in \mathcal{K}$ indexes the diagnostic circuit routes defined in Table~\ref{app-tab:circuit_presets}.
We compare each route against the unsteered base model:
\begin{equation}
s_k(x,\tilde{y}) = \ell_k(\tilde{y}\mid x) -\ell_0(\tilde{y}\mid x),
\end{equation}
where $\ell_0$ denotes the average token log-likelihood under the unsteered model. 
The resulting support score $s_k(x,\tilde{y})$ quantifies how much route $k$ increases the candidate's likelihood relative to the unsteered model, providing an internal compatibility check for the intended circuit route.

As illustrated in Figure~\ref{app-fig:selction}, this mechanistic filtering effectively shifts the distribution of generated data toward the desired utility profiles. The selector therefore serves as a lightweight consistency check on top of steering, retaining outputs that are both surface-valid and internally aligned with the target generation objective.

\subsection{Circuit-Steered Generation Hyperparameters}
As introduced in Section~\ref{circuit_steering}, we construct three targeted data pools, $\mathcal{D}_{\text{learnable}}$, $\mathcal{D}_{\text{challenging}}$, and $\mathcal{D}_{\text{aligned}}$.

\begin{table}[h]
\centering
\small
\caption{Common generation settings used across all utility profiles for both circuit-steered and prompt-based baseline generations.}
\label{app-tab:common_hparams}
\vspace{0.2em}
\begin{tabular}{ll}
\toprule
\textbf{Hyperparameter} & \textbf{Value} \\
\midrule
Base model & Qwen2.5-1.5B-Instruct \\
Seed examples & 5,000 SciQ training examples \\
Candidates per source example & 4 \\
Selected candidates per source example & 2 \\
Sampling temperature & 0.8 \\
Top-$p$ & 0.95 \\
Maximum new tokens & 256 \\
Seed & 42 \\
\bottomrule
\end{tabular}
\end{table}

\paragraph{Generation Configurations.} 
To strictly isolate the effect of mechanistic steering from superficial prompt-following, we use a single, target-independent fixed prompt alongside the common generation configurations (Table~\ref{app-tab:common_hparams}) for all circuit-steered generation runs. By keeping the textual instructions entirely neutral, we force the model to rely exclusively on the steering mechanisms to induce the desired utility profiles. As a comparative baseline, we also evaluate standard prompt-based generation. For this baseline, we augment the original source questions with explicit textual instructions that request learnable(easier), challenging(harder), or more aligned(instructional) completions, without applying any internal intervention. Each prompt includes three few-shot demonstrations to provide the desired response format and target style. The overarching prompt templates and the specific textual injections used for each setting are detailed in Table~\ref{app-tab:prompt_template}.

\begin{table*}[h]
\centering
\small
\caption{Prompt templates for generation. Circuit-steered generation uses the same target-independent fixed prompt for all profiles, whereas prompt-based baselines use explicit profile-specific instructions.}
\label{app-tab:prompt_template}
\begin{tabularx}{\textwidth}{l X}
\toprule
\textbf{Setting} & \textbf{Prompt Template} \\
\midrule
Shared system instruction &
Generate science multiple-choice training data. Keep the sample factually correct, avoid trivia leakage, and return strict JSON. Each choice must be plain text without letter or number prefixes. The answer field must be exactly one uppercase letter among \texttt{A}, \texttt{B}, \texttt{C}, and \texttt{D}. \\
\midrule
Circuit-steered fixed prompt &
\textbf{Source topic:} \texttt{\{topic\}} \newline
\textbf{Source item:} \texttt{\{source question\}} \newline
\textbf{Source answer:} \texttt{\{optional source answer\}} \newline
Create one new science MCQA sample in a similar domain.
Keep it self-contained, factually correct, and naturally written.
Do not explicitly mention a target difficulty band in the output.
Return only JSON with fields
\texttt{question}, \texttt{choices}, \texttt{answer}, \texttt{rationale}, \texttt{difficulty}, and \texttt{helpful}. \\
\midrule
Prompt baseline: learnable-style &
Create one new MCQA sample that is easier than the source.
Use a single canonical science fact or one direct concept application.
Distractors should be plausible but clearly eliminable by the decisive concept.
Return only JSON with the same schema. \\
\midrule
Prompt baseline: challenging-style &
Create one new MCQA sample that is harder than the source.
Require at least one extra inference step, concept disambiguation, or stronger distractor discrimination.
Return only JSON with the same schema. \\
\midrule
Prompt baseline: aligned-style &
Create one new MCQA sample optimized to be instructional and helpful.
The rationale must state the decisive evidence and why at least one distractor is wrong.
Return only JSON with the same schema. \\
\bottomrule
\end{tabularx}
\end{table*}

\paragraph{Circuit Presets and Steering Scales.}
To perform the interventions, we extract the most critical steering targets from the EAP-IG circuit analysis by selecting nodes based on their aggregated edge attribution mass, as summarized in Table~\ref{app-tab:circuit_presets}. These preset positions coincide with the dominant information-processing pathways identified in our structural analysis (Table~\ref{app-tab:struct_circuits}), and are therefore used as the primary targets for circuit steering. We then apply profile-specific continuous scaling factors for both activation addition and attention modulation. The complete hyperparameter configurations for mapping each utility profile to its corresponding internal intervention scales are detailed in Table~\ref{app-tab:profile-scales}.

\begin{table*}[h]
\centering
\small
\setlength{\tabcolsep}{5pt}
\renewcommand{\arraystretch}{1.15}
\caption{Default circuit steering presets used for Qwen2.5-1.5B-Instruct on SciQ. All listed locations are used as activation-addition targets when constructing utility-specific steering directions. Targets marked with $\star$ additionally receive targeted attention-logit steering during decoding.
}
\label{app-tab:circuit_presets}
\begin{tabularx}{\textwidth}{l l X}
\toprule
\textbf{Circuit Preset} & \textbf{Functional Role} & \textbf{Steering Targets} \\
\midrule
AUM-High &
Canonical anchor &
$\mathrm{a0.h[6\!-\!11]}\langle V\rangle^{\star}$;
$m_1$;
$m_2$;
$\mathrm{a6.h[6\!-\!11]}\langle K\rangle^{\star}$;
$\mathrm{a14.h2}\langle K\rangle^{\star}$;
$m_{23}$ \\
AUM-Low &
Boundary branch &
$\mathrm{a2}\langle K\rangle$;
$m_3$;
$\mathrm{a5}\langle K\rangle$;
$m_{27}$ \\
\midrule
EL2N-Low &
Low-pressure anchor &
$\mathrm{a0.h[6\!-\!11]}\langle V\rangle^{\star}$;
$m_1$;
$m_2$;
$m_3$;
$\mathrm{a6.h[6\!-\!11]}\langle K\rangle^{\star}$;
$\mathrm{a7}\langle K\rangle$;
$m_{23}$ \\
EL2N-High &
Informative rerouting &
$\mathrm{a2}\langle K\rangle$;
$\mathrm{a4}\langle K\rangle$;
$\mathrm{a5}\langle K\rangle$;
$m_4$;
$m_6$;
$m_{13}$ \\
\midrule
GradAlign-High &
Target-aligned control &
$m_{13}$;
$\mathrm{a14.h2}\langle K\rangle^{\star}$;
$\mathrm{a17}$--$\mathrm{a20}$ late-control attention targets \\
GradAlign-Low &
Shortcut evidence &
$\mathrm{a0.h[6\!-\!11]}\langle V\rangle^{\star}$;
$m_1$;
$m_2$;
$m_3$;
$m_{23}$ \\
\bottomrule
\end{tabularx}
\end{table*}

\begin{table}[t]
\centering
\small 
\setlength{\tabcolsep}{2.7pt} 
\caption{Per-profile activation and attention steering scales applied to each circuit preset. Values are presented as \textbf{Activation Scale (Attention Scale)}. Positive values amplify the corresponding circuit's forward-pass contribution, while negative values suppress it. GradAlign-H inherently targets $\text{a14}\langle K\rangle$ with an inverted weight, thus, a positive recipe scale yields suppressive steering on that specific node.}
\label{app-tab:profile-scales}
\vspace{0.3em}
\begin{tabular}{lcccccc}
\toprule
& \multicolumn{6}{c}{\textbf{Activation Scale (Attention Scale)}} \\
\cmidrule(lr){2-7}
\textbf{Profile} & \textbf{AUM-H} & \textbf{AUM-L} & \textbf{EL2N-H} & \textbf{EL2N-L} & \textbf{GradAlign-H} & \textbf{GradAlign-L} \\
\midrule
$\mathcal{D}_{\text{learnable}}$
  & $+0.40$ ($+0.28$) & $-0.25$ ($0.00$) & $-0.35$ ($-0.14$) & $+0.47$ ($+0.31$) & $+0.10$ ($+0.02$) & $+0.12$ ($+0.12$) \\
$\mathcal{D}_{\text{challenging}}$
  & $-0.37$ ($-0.18$) & $0.00$ ($0.00$) & $+0.40$ ($+0.15$) & $-0.40$ ($-0.21$) & $+0.36$ ($+0.15$) & $-0.30$ ($-0.18$) \\
$\mathcal{D}_{\text{aligned}}$
  & $+0.22$ ($+0.10$) & $-0.14$ ($0.00$) & $-0.10$ ($-0.06$) & $+0.20$ ($+0.10$) & $+0.45$ ($+0.17$) & $-0.28$ ($-0.15$) \\
\bottomrule
\end{tabular}
\end{table}

\section{Generated Dataset Statistics}
\label{app:gen_statistics}

\paragraph{Basic statistics.}
We summarize the basic statistics of the resulting generated datasets in Table~\ref{app-tab:gen_statistics}. From an initial pool of 20,000 generation attempts per profile, we selected the top 10,000 valid samples based on our internal compatibility selection. 

\begin{table*}[h]
\centering
\caption{Basic statistics of generated datasets after selection. JSON denotes the percentage of candidates successfully parsed into the required MCQA JSON format. BLEU-1 and Jaccard are computed between the source question and the generated question. Lower values indicate less lexical copying from the seed sample. 
}
\label{app-tab:gen_statistics}
\setlength{\tabcolsep}{5pt}
\begin{tabular}{lrrrrrrr}
\toprule
\textbf{Profile} &
\textbf{Raw} &
\textbf{Selected} &
\textbf{JSON} &
\textbf{Q. Len.} &
\textbf{Rat. Len.} &
\textbf{BLEU-1} &
\textbf{Jaccard} \\
\midrule
Circuit-learnable    & 20{,}000 & 10{,}000 & 98.2 & 9.9  & 20.0 & 0.2752 & 0.1507 \\
Circuit-challenging  & 20{,}000 & 10{,}000 & 99.5 & 10.8 & 22.6 & 0.2495 & 0.1395 \\
Circuit-aligned      & 20{,}000 & 10{,}000 & 99.2 & 10.1 & 20.5 & 0.2680 & 0.1468 \\
\midrule
Prompt-learnable     & 20{,}000 & 10{,}000 & 99.5 & 9.5  & 20.7 & 0.3150 & 0.1725 \\
Prompt-challenging   & 20{,}000 & 10{,}000 & 98.7 & 12.9 & 36.6 & 0.2473 & 0.1500 \\
Prompt-aligned       & 20{,}000 & 10{,}000 & 98.5 & 10.3 & 40.6 & 0.4039 & 0.2497 \\
\bottomrule
\end{tabular}
\end{table*}

The JSON parsing rate is consistently high across both circuit-steered and prompt-based generation, indicating that the generation pipeline reliably follows the required structured MCQA format. Additionally, as expected, $\mathcal{D}_{\text{challenging}}$ exhibits slightly longer questions and rationales compared to $\mathcal{D}_{\text{learnable}}$, reflecting the increased complexity of the underlying reasoning tasks. Interestingly, the prompt-based baselines yield significantly longer rationales. Qualitative inspection indicates that this excess length in prompt-based generation is largely driven by conversational fillers and stylistic verbosity (e.g., "This problem requires..."), rather than dense, informative reasoning.

Furthermore, we evaluate the lexical diversity of the generated questions by measuring their overlap with the original seed questions using BLEU-1 and Jaccard similarity. The circuit-steered profiles demonstrate notably lower similarity scores than the prompt-based baselines. This indicates that our mechanistic steering successfully induces diverse internal generation trajectories, relying less on superficial copying of the source text and more on fundamental concept manipulation.

\paragraph{Internal proxy alignment.} 
To verify that the generated samples align with their intended circuit profiles without relying solely on human evaluation, we introduce reference-free internal diagnostic metrics termed \emph{Proxy Measures}. We evaluate how the average completion log-likelihood shifts when specific circuit profiles are artificially activated. 

Recall from Section~\ref{app:gen_details} that the circuit-specific support score is defined as $s_k(x, \tilde{y}) = \ell_k(\tilde{y} \mid x) - \ell_0(\tilde{y} \mid x)$. A positive value indicates that the target circuit intrinsically prefers the generated completion. Using these support values, we define a steering diagnostic as Equation~\eqref{app-eq:diff_proxy}. A high \texttt{Difficulty Proxy} indicates the completion leans toward challenging reasoning rather than canonical anchors. 
\begin{equation}
\label{app-eq:diff_proxy}
    \text{Difficulty Proxy} = s_{\text{challenging}}(x, \tilde{y}) - s_{\text{learnable}}(x, \tilde{y}) 
\end{equation}

\begin{figure}[h]
  \centering
  \includegraphics[trim=0cm 5cm 0cm 3.5cm, clip, width=\columnwidth]{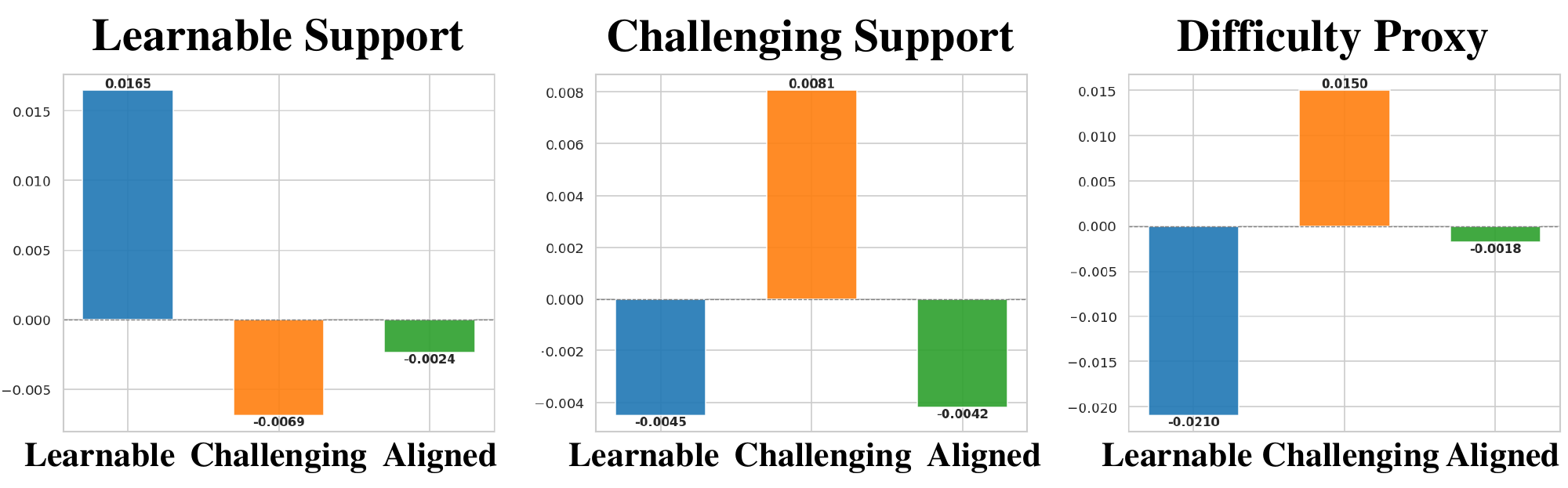}
  \caption{Internal proxy alignment across the generated datasets. The distributions of support scores and the difficulty proxy show that each dataset successfully aligns with its targeted circuit profile.}
  \label{app-fig:proxy}
\end{figure}

As illustrated in Figure~\ref{app-fig:proxy}, the proxy distributions shift precisely as intended. $\mathcal{D}_{\text{learnable}}$ skews toward lower difficulty proxies, $\mathcal{D}_{\text{challenging}}$ robustly shifts toward higher difficulty proxies, and $\mathcal{D}_{\text{aligned}}$ maintains a well-balanced, moderate profile. This confirms that our framework successfully aligns the generated outputs with the targeted internal processing mechanisms.

\section{Generation Fidelity Experiments Details.} 
\label{app:gen_fid_details}
\paragraph{Evaluation Metrics.}
We evaluate generated datasets along three criteria, complexity, diversity, and quality, following prior work~\cite{dataselQDC,datadiversityit,surveyQDC}. Below, we describe the metrics used for the main comparison reported in Figure~\ref{fig:3axisradar}.

\textit{Complexity} characterizes how challenging the generated samples are for the proxy model. \textbf{PPL} (mean perplexity of the proxy model on generated text) reflects the structural difficulty and lexical unpredictability of the samples, where higher values indicate greater challenge. \textbf{Margin} (the logit gap between the correct and highest-scoring distractor answer, evaluated without rationale) measures how discriminable the correct answer is. Lower margin indicates a harder, more ambiguous question.

\textit{Diversity} captures distributional breadth of the generated set. \textbf{Vendi} is computed over sentence-level embeddings(\texttt{all-MiniLM-L6-v2}) and measures lexical variety. \textbf{G-Vendi} replaces text embeddings with per-sample input-gradient fingerprints from the proxy model, measuring diversity from a training-dynamics perspective, how differently each sample perturbs the model's parameters.

\textit{Quality} assesses whether the generated samples constitute valid learning signal. \textbf{Ans-Acc} measures proxy model accuracy on questions without rationale, how well the proxy model can answer the generated questions. \textbf{R-Gain} is the increase in the proxy model's correct-answer probability when the rationale is provided,  measuring how informative the rationale is. Higher R-Gain indicates questions that are well-structured yet genuinely supported by the accompanying rationale.

\paragraph{Utility-axis Ablation and Diversity Score Computation.}
For the ablation study in Section~\ref{gen_fidelity}, we construct three leave-one-out baselines (\emph{w/o AUM}, \emph{w/o EL2N}, \emph{w/o GradAlign}) by setting the activation and attention steering scales of the corresponding axis to zero, while keeping all other conditions identical to the \emph{Full} configuration (Table~\ref{app-tab:profile-scales}). This isolates the contribution of each circuit axis to the overall data quality. Vendi is computed over up to $1{,}000$ samples per condition embedded with \texttt{sentence-transformers/all-MiniLM-L6-v2}. G-Vendi is computed over up to 300 samples by extracting $\ell_2$-normalized input-gradient fingerprints from \texttt{Qwen/Qwen2.5-0.5B-Instruct} and applying the Vendi score to them. \textbf{Recall} is estimated against up to $1{,}800$ reference samples from the SciQ training split using $k{=}5$ nearest neighbors in embedding space. 

\section{SAMS Downstream Fine-tuning Details}
\label{app:downstream_details}
This section provides additional implementation details for the downstream fine-tuning experiments in Section~\ref{overall_sams}. All downstream experiments use \texttt{Qwen/Qwen2.5-0.5B-Instruct} as the student model.

\paragraph{Data mixture construction.}
For each source ratio $\rho \in \{50\%, 60\%, 70\%\}$, we construct each training batch by sampling a fraction $\rho$ from $\mathcal{D}_{\mathrm{src}}$ and the remaining fraction $1-\rho$ from generated data. Thus, all methods are compared under the same total training budget, and the source ratio controls only the relative amount of original versus synthetic data. For the \textsc{Uniform-Mix} setting, the generated portion of each batch is constructed by sampling uniformly from each of the three synthetic pools with equal probability:
$$ x_{\mathrm{syn}} \sim \frac{1}{3} \mathrm{Unif}(\mathcal{D}_{\mathrm{learnable}}) + \frac{1}{3} \mathrm{Unif}(\mathcal{D}_{\mathrm{challenging}}) + \frac{1}{3} \mathrm{Unif}(\mathcal{D}_{\mathrm{aligned}}) $$
where $\mathrm{Unif}(\mathcal{D})$ denotes a uniform sampling distribution over the given dataset. For SAMS, the generated portion is sampled according to the stage-specific curriculum described below.

\paragraph{Stage-aware mechanistic scheduling.}
SAMS divides fine-tuning into three stages: warm-up, transition, and challenge. Unless otherwise specified, we divide the total training steps into three equal intervals and apply the corresponding synthetic-pool ratios in Algorithm~\ref{alg:mechanistic_scheduling}. Specifically, the aligned pool is kept active throughout training, while the sampling mass gradually shifts from the learnable pool to the challenging pool. This schedule preserves early optimization stability while increasing structurally demanding examples in later stages.

\paragraph{Fine-tuning hyperparameters.}
All downstream fine-tuning experiments use the same optimization hyperparameters, summarized in Table~\ref{app-tab:downstream_hparams}. We train for six epochs with AdamW, using a learning rate of $2\times10^{-5}$ on an RTX Pro 6000 GPU.

\begin{table}[h]
\centering
\caption{Fine-tuning hyperparameters for downstream validation experiments. The same configuration is used for all methods, schedules, and source ratios.}
\label{app-tab:downstream_hparams}
\vspace{0.2em}
\begin{tabular}{ll}
\toprule
\textbf{Hyperparameter} & \textbf{Value} \\
\midrule
Training epochs & 6 \\
Per-device batch size & 12 \\
Gradient accumulation steps & 4 \\
Effective batch size & 48 \\
Evaluation batch size & 8 \\
Learning rate & $2\times10^{-5}$ \\
Weight decay & 0.01 \\
Warm-up ratio & 0.03 \\
Source ratios & 50\%, 60\%, 70\% \\
\bottomrule
\end{tabular}
\end{table}

\paragraph{Evaluation protocol.}
We evaluate each fine-tuned model on SciQ test set as the in-domain benchmark and ARC-Easy test set as the out-of-domain benchmark. For both datasets, we report accuracy, expected calibration error (ECE), and negative log-likelihood (NLL). Accuracy measures task performance, while ECE and NLL capture complementary aspects of probabilistic calibration. This evaluation protocol allows us to test not only whether the generated data improves in-domain performance, but also whether mechanistically scheduled synthetic data supports more reliable out-of-domain generalization.

\paragraph{Stronger prompt-level baselines.} 
Table~\ref{app-tab:prompt_template} provides the full prompt templates, which explicitly instruct the generator to produce learnable, helpful, and challenging examples, making the prompt baseline substantially stronger than generic prompting. Nevertheless, we further add stronger prompt and circuit baselines, including utility-specific rejection sampling.

\begin{table*}[h]
    \centering
    \caption{Comparison with stronger prompt and circuit baselines.}
    \label{tab:strong_baselines}
    \small
    \setlength{\tabcolsep}{5pt}
    \begin{tabular}{lcccc}
        \toprule
        Config & SciQ ACC $\uparrow$ & SciQ ECE $\downarrow$ & SciQ NII $\downarrow$ & ARC (OOD) $\uparrow$ \\
        \midrule
        Prompt-Uniform
            & $83.87 \pm 0.38$ & $0.077 \pm 0.006$ & $0.543 \pm 0.015$ & $73.79 \pm 0.21$ \\
        Prompt + Diversity
            & $83.30 \pm 0.28$ & $0.085 \pm 0.021$ & $0.555 \pm 0.064$ & $72.16 \pm 0.45$ \\
        Prompt + Reject-AUM
            & $84.10 \pm 0.79$ & $0.053 \pm 0.040$ & $0.490 \pm 0.095$ & $76.01 \pm 0.99$ \\
        Prompt + Reject-EL2N
            & $82.20 \pm 0.99$ & $0.080 \pm 0.028$ & $0.555 \pm 0.078$ & $64.98 \pm 6.85$ \\
        Prompt + Reject-GA
            & $83.43 \pm 0.60$ & $0.043 \pm 0.012$ & $0.477 \pm 0.006$ & $73.71 \pm 0.28$ \\
        Random-Circuit steering
            & $83.93 \pm 0.50$ & \textbf{$0.040 \pm 0.010$} & \textbf{$0.460 \pm 0.017$} & $74.26 \pm 0.84$ \\
        Circuit-Uniform
            & $84.77 \pm 0.38$ & $0.063 \pm 0.006$ & $0.510 \pm 0.026$ & $76.09 \pm 0.25$ \\
        SAMS (ours)
            & \textbf{$85.20 \pm 0.56$} & $0.057 \pm 0.023$ & $0.490 \pm 0.053$ & \textbf{$76.15 \pm 0.30$} \\
        \bottomrule
    \end{tabular}
\end{table*}

SAMS remains best on both accuracy metrics against every prompt-level competitor. Among prompt-only methods, Reject-AUM is the strongest baseline but still falls short of SAMS.

\section{Limitations and Future Work}
\label{app:limitation}
This work establishes a transparent, mechanistic link between internal information processing and sample-level training utility. By moving beyond black-box data metrics, we demonstrate that microscopic circuit-level observations can be successfully leveraged for practical macroscopic applications, such as targeted data filtering and circuit-steered generation.

One natural extension of our current framework is to make the SAMS scheduling policy more systematic. In this work, SAMS uses a stage-wise mixture inspired by curriculum learning principles~\cite{curri}, reflecting the intuition that different phases of fine-tuning benefit from different utility profiles. While effective, this design does not yet feature a fully automated scheduling policy. Nevertheless, the empirical gains from SAMS support our central insight that circuit-steered data are most useful when their utility profiles are aligned with the model's evolving optimization needs. Future work can build on this direction by exploring a broader family of stage-wise ratios and learning scheduling policies that adjust the mixture during training. Ultimately, we view this work as an initial step toward white-box, mechanistically guided data curation pipelines that can not only analyze training data quality, but actively shape it according to the model's internal learning dynamics.


\clearpage
\input{checklist.tex}

\end{document}

%% file: checklist.tex
\section*{NeurIPS Paper Checklist}

\begin{enumerate}

\item {\bf Claims}
    \item[] Question: Do the main claims made in the abstract and introduction accurately reflect the paper's contributions and scope?
    \item[] Answer: \answerYes{} 
    \item[] Justification: The abstract and introduction convey the paper's key contribution of bridging mechanistic interpretability and data generation. They are well aligned with the proposed methods and experimental scope.

\item {\bf Limitations}
    \item[] Question: Does the paper discuss the limitations of the work performed by the authors?
    \item[] Answer: \answerYes{} 
    \item[] Justification: The paper discusses limitations and future work directions at the end of the conclusion (\ref{limitation}), with additional details provided in the Appendix~\ref{app:limitation}. 

\item {\bf Theory assumptions and proofs}
    \item[] Question: For each theoretical result, does the paper provide the full set of assumptions and a complete (and correct) proof?
    \item[] Answer: \answerNA{} 
    \item[] Justification: The paper does not include theoretical results.

    \item {\bf Experimental result reproducibility}
    \item[] Question: Does the paper fully disclose all the information needed to reproduce the main experimental results of the paper to the extent that it affects the main claims and/or conclusions of the paper (regardless of whether the code and data are provided or not)?
    \item[] Answer: \answerYes{} 
    \item[] Justification:  We include detailed experimental setups for reproducibility in both the main section (\ref{circuit_setup},\ref{circuit_steering},\ref{overall_sams}) and in the appendix (\ref{app:discovery_details}, \ref{app:filtering_details}, \ref{app:gen_details}).

\item {\bf Open access to data and code}
    \item[] Question: Does the paper provide open access to the data and code, with sufficient instructions to faithfully reproduce the main experimental results, as described in supplemental material?
    \item[] Answer: \answerYes{} 
    \item[] Justification: Our source code is included in the supplementary material. A representative subset of the data can be found within the \texttt{datas/} folder of the provided codebase.

\item {\bf Experimental setting/details}
    \item[] Question: Does the paper specify all the training and test details (e.g., data splits, hyperparameters, how they were chosen, type of optimizer) necessary to understand the results?
    \item[] Answer: \answerYes{} 
    \item[] Justification: The paper provides detailed experimental settings in Appendix~\ref{app:discovery_details}, Appendix~\ref{app:filtering_details}, Appendix~\ref{app:gen_details},  and Appendix~\ref{app:gen_fid_details}. These sections cover the necessary setup details, implementation choices, and hyperparameters, to understand the reported results.

\item {\bf Experiment statistical significance}
    \item[] Question: Does the paper report error bars suitably and correctly defined or other appropriate information about the statistical significance of the experiments?
    \item[] Answer: \answerYes{} 
    \item[] Justification: We report error bars where repeated stochastic trials are meaningful and feasible, such as in Figure~\ref{fig:gen_fidelity}. For deterministic circuit-level analyses, we provide causal intervention results and random-circuit baselines to support the reliability of the findings.

\item {\bf Experiments compute resources}
    \item[] Question: For each experiment, does the paper provide sufficient information on the computer resources (type of compute workers, memory, time of execution) needed to reproduce the experiments?
    \item[] Answer: \answerYes{} 
    \item[] Justification: The required compute resource information is reported alongside the experimental settings in the corresponding appendix sections. 
    
\item {\bf Code of ethics}
    \item[] Question: Does the research conducted in the paper conform, in every respect, with the NeurIPS Code of Ethics \url{https://neurips.cc/public/EthicsGuidelines}?
    \item[] Answer: \answerYes{} 
    \item[] Justification: We have reviewed the NeurIPS Code of Ethics throughout this work.

\item {\bf Broader impacts}
    \item[] Question: Does the paper discuss both potential positive societal impacts and negative societal impacts of the work performed?
    \item[] Answer: \answerNA{} 
    \item[] Justification: This work is primarily methodological and does not involve real-world deployment, human-subject studies, or applications targeting specific social groups. Thus, direct positive or negative societal impacts are not applicable beyond the general implications for model transparency and controllability.
    
\item {\bf Safeguards}
    \item[] Question: Does the paper describe safeguards that have been put in place for responsible release of data or models that have a high risk for misuse (e.g., pre-trained language models, image generators, or scraped datasets)?
    \item[] Answer: \answerNA{} 
    \item[] Justification: This work does not release high-risk artifacts such as pretrained models, scraped datasets, or user-sensitive data. The released code and representative small-scale samples are for research reproducibility and do not introduce special misuse risks requiring additional safeguards.

\item {\bf Licenses for existing assets}
    \item[] Question: Are the creators or original owners of assets (e.g., code, data, models), used in the paper, properly credited and are the license and terms of use explicitly mentioned and properly respected?
    \item[] Answer: \answerYes{} 
    \item[] Justification: We acknowledge all external assets used in this work by citing either their original URLs or the relevant publications.

\item {\bf New assets}
    \item[] Question: Are new assets introduced in the paper well documented and is the documentation provided alongside the assets?
    \item[] Answer: \answerYes{} 
    \item[] Justification: The newly introduced assets are documented in Appendix~\ref{app:gen_statistics}, including dataset statistics and construction details.

\item {\bf Crowdsourcing and research with human subjects}
    \item[] Question: For crowdsourcing experiments and research with human subjects, does the paper include the full text of instructions given to participants and screenshots, if applicable, as well as details about compensation (if any)? 
    \item[] Answer: \answerNA{} 
    \item[] Justification: The paper does not involve crowdsourcing nor research with human subjects.

\item {\bf Institutional review board (IRB) approvals or equivalent for research with human subjects}
    \item[] Question: Does the paper describe potential risks incurred by study participants, whether such risks were disclosed to the subjects, and whether Institutional Review Board (IRB) approvals (or an equivalent approval/review based on the requirements of your country or institution) were obtained?
    \item[] Answer: \answerNA{} 
    \item[] Justification: The work does not involve crowdsourcing or research conducted with human subjects.

\item {\bf Declaration of LLM usage}
    \item[] Question: Does the paper describe the usage of LLMs if it is an important, original, or non-standard component of the core methods in this research? Note that if the LLM is used only for writing, editing, or formatting purposes and does \emph{not} impact the core methodology, scientific rigor, or originality of the research, declaration is not required.
    \item[] Answer: \answerNA{} 
    \item[] Justification: LLMs were solely used as writing and formatting assistants. They were not used as an important component of the proposed method, analysis, or conclusions.

\end{enumerate}